\documentclass[sigconf]{acmart}
\usepackage{algorithm}
\usepackage{algorithmic}
\usepackage{balance}
\usepackage{newfloat}
\usepackage{listings}
\usepackage{nicematrix}
\usepackage{multirow}
\usepackage{cleveref}
\usepackage{booktabs}
\usepackage{rotating}
\usepackage[inline]{enumitem}
\usepackage{tabularray}
\usepackage{array}
\usepackage{xcolor}
\usepackage{makecell}
\usepackage{mdframed,framed,lipsum}
\usepackage[misc]{ifsym}
\usepackage{soul}
\usepackage{pifont}
\usepackage{threeparttable}
\AtBeginDocument{%
  }

\copyrightyear{2026}
\acmYear{2026}
\acmConference[WWW '26]{Proceedings of the ACM Web Conference 2026}{April 13--17, 2026}{Dubai, United Arab Emirates}

\def\eg{\emph{e.g.}} 
\def\ie{\emph{i.e.}} 

\def\etc{\emph{etc.}}

\def\qwenvl{\textbf{Qwen2-VL-7B}}
\def\qwenllm{\textbf{Qwen2.5-14B}}
\definecolor{mygray}{RGB}{192,192,192}
\definecolor{mygreen}{RGB}{214,228,210}
\def\Ignore{\textcolor[RGB]{128, 128, 128}{[Ignore]}}
\def\Outformat{\textcolor[RGB]{46,139,87}{[Fm]}}
\def\ocr{\textcolor[RGB]{60,179,113}{[OCR]}}
\def\VIG{\textcolor[RGB]{138,43,226}{[VIG]}}
\def\MtoT{\textcolor[RGB]{255, 140, 0}{[M2T]}}
\def\fewshot{\textcolor[RGB]{255,105,180}{[FS]}}
\def\cot{\textcolor[RGB]{0,191,255}{[CoT]}}
\def\GL{\textcolor[RGB]{255,0,0}{<GL>}}

\begin{document}

\title{Read as You See: Guiding Unimodal LLMs for Low-Resource Explainable Harmful Meme Detection}

\author{Fengjun Pan}
\orcid{0000-0002-2470-1958}
\affiliation{%
  \institution{Nanyang Technological University}
  \city{Singapore}
  \country{Singapore}}
\email{fengjun001@e.ntu.edu.sg}

\author{Xiaobao Wu}
\authornote{Corresponding Author.}
\orcid{0000-0003-0076-3924}
\affiliation{%
 \institution{Shanghai Jiao Tong University}
 \city{Shanghai}
 \country{China}}
\email{xiaobaowu@sjtu.edu.cn}

\author{Tho Quan}
\orcid{0000-0003-0467-6254}
\affiliation{%
 \institution{Ho Chi Minh City University of Technology}
 \city{Ho Chi Minh City}
 \country{Vietnam}}
\email{qttho@hcmut.edu.vn}

\author{Anh Tuan Luu}
\authornotemark[1]
\orcid{0000-0001-6062-207X}
\affiliation{%
  \institution{Nanyang Technological University}
  \city{Singapore}
  \country{Singapore}}
\affiliation{%
  \institution{VinUniversity}
  \city{Hanoi}
  \country{Vietnam}}
\email{anhtuan.luu@ntu.edu.sg}
\renewcommand{\shortauthors}{Fengjun Pan, Xiaobao Wu, Tho Quan, and Anh Tuan Luu}
\begin{abstract}
    Detecting harmful memes is crucial for safeguarding the integrity and harmony of online environments, yet existing detection methods are often resource-intensive, inflexible, and lacking explainability,
    limiting their applicability in assisting real-world web content moderation.
    We propose \textbf{U-CoT+}, a resource-efficient framework that prioritizes accessibility, flexibility and transparency in harmful meme detection by fully harnessing the capabilities of lightweight unimodal large language models (LLMs).
    Instead of directly prompting or fine-tuning large multimodal models (LMMs) as black-box classifiers, we avoid immediate reasoning over complex visual inputs but decouple meme content recognition from meme harmfulness analysis through a high-fidelity meme-to-text pipeline, which collaborates lightweight LMMs and LLMs to convert multimodal memes into natural language descriptions that preserve critical visual information, thus enabling text-only LLMs to ``see'' memes by ``reading''.
    Grounded in textual inputs, we further guide unimodal LLMs' reasoning under zero-shot Chain-of-Thoughts (CoT) prompting with targeted, interpretable, context-aware, and easily obtained human-crafted guidelines, thus providing accountable step-by-step rationales, while enabling flexible and efficient adaptation to diverse sociocultural criteria of harmfulness.
    Extensive experiments on seven benchmark datasets show that U-CoT+ achieves performance comparable to resource-intensive baselines,
    highlighting its effectiveness and potential as a scalable, explainable, and low-resource solution to support harmful meme detection\footnote{Codes and data are available at: {\hypersetup{urlcolor=magenta}\url{https://github.com/panFJCharlotte98/HMC}}}.
\end{abstract}


\maketitle

\section{Introduction}
    As a popular communication medium on social media platforms~\cite{meme_survey,memes_socialmediaplatform},
    memes convey rich information in a visually engaging manner by combining images with overlaid text~\cite{internet_memes}. 
    Despite their often amusing nature, memes can be exploited to disseminate harmful content under the guise of humor, such as hate speech~\cite{FHM_kiela2020hateful,davidson2019racial}, toxic misinformation~\cite{lu2024towards,kennedy2022repeat},
    and inflammatory opinions~\cite{liu2022figmemes,edwards2013pathways}.
    Harmful memes\footnote{\textcolor[rgb]{1,0,0}{Disclaimer: This paper contains demonstration content that may be disturbing.}}
    pose a threat to the safety and harmony of online environments~\cite{ToxicMemesSurvey_pandiani2024toxic,sharma2022detecting},
    thus motivating automated harmful meme detection to support content moderation for online platforms.
    
    However, existing methods, most of which follow a typical supervised fine-tuning (SFT) paradigm, face three critical limitations:
    \begin{enumerate*}[label=\textbf{(\roman*)}]
        \item
            \textbf{Poor Resource Efficiency}.
            On one hand, supervised approaches tend to be data-inefficient as they require large volumes of labeled training data to perform well~\cite{ISSUES_burbi2023mapping, PrideMM_shah2024memeclip,HateClipper_kumar2022hate,HarmP_pramanick2021momenta}. 
            However, the nuanced, implicit and context-specific nature of memes~\cite{chakravarthi2025findings,baruah2020context} makes it difficult for human annotators to maintain consistent labeling without being limited by individual knowledge and subjectivity~\cite{gillespie2020content}.
            As such, collecting and annotating memes are often time-consuming, costly, and prone to bias.
            On the other hand, existing low-resource methods~\cite{LowResource_huang2024towards,CapAlign_ji2024capalign,ExplainHM_lin2024towards} often rely on proprietary Large Multimodal Models (LMMs) \eg, GPT-4 or heavily scaled models with massive parameter counts,
            incurring substantial financial and computational expenses.
            Taken together, these methods pose significant accessibility challenges for practitioners operating under low-resource constraints.
        \item
            \textbf{Limited Scalability and Flexibility}.
            The classification criteria distinguishing harmful and harmless memes often vary across sociocultural contexts~\cite{zannettou2018origins,sharma2022detecting} and differ across platforms, regions, and over time~\cite{bonagiri2025towards,singhal2023sok,pater2016characterizations,gorwa2020algorithmic}, which entails the use of specialized policies for different scenarios in order to robustly define harmful memes~\cite{thomas2025supporting}.
            However, supervised classifiers are not easily scalable. 
            Once trained, they become difficult to adapt to new out-of-domain criteria, meme or harmfulness types without undergoing costly data collection and retraining~\cite{RGCLnew_mei2025improved,modhate}.
        \item
            \textbf{Lack of Explainability}.
            Real-world content moderation practices combine model-generated decisions to assist human verdicts~\cite{bonagiri2025towards,FacebookHelp2024}. 
            However, most supervised methods operate as black boxes, offering no human-interpretable reasoning behind their predictions~\cite{burkart2021survey,Uncertainty_yang2024uncertainty,ISSUES_burbi2023mapping}.
            This lack of transparency in decision-making undermines trust in human-machine collaboration and deployment reliability in sensitive or regulated scenarios~\cite{sasikala2024decoding},
            where explicit rationales are essential to support subsequent human reviews and feedback.
    \end{enumerate*}

    To address these limitations, in this paper, we explore the feasibility of harnessing lightweight Large Language Models (LLMs) for low-resource harmful meme detection.
    Specifically, we propose \textbf{U-CoT+}, a unimodal guided Chain-of-Thoughts (CoT)~\cite{wei2022chain} prompting framework, which converts the default multimodal setup back to unimodal text-only inference, enabling LLMs to perform zero-shot harmful meme detection solely from textual input.
    To this end, we first propose a \textbf{High-Fidelity Meme2Text} pipeline.
    Instead of directly prompting LMMs to perform classification,
    we query lightweight LMMs multiple times to extract critical visual cues~\cite{Procap_cao2023pro,CapAlign_ji2024capalign} for sensitive, identity-related attributes of human subjects (\eg, race, gender).
    We then resort to a unimodal LLM to integrate the LMM outputs into a unified, informative description which it subsequently uses for detection.
    This design effectively decouples content recognition from harmfulness analysis,
    reducing task complexity by avoiding immediate reasoning over complicated visual elements.
    
    Furthermore, we introduce the \textbf{Guided CoT} approach.
    Rather than relying solely on LLMs' internalized knowledge, 
    we guide them with explicit, human-crafted guidelines that specify fine-grained, targeted criteria for distinguishing harmful from harmless memes.
    LLMs reason step by step over textual meme descriptions in conjunction with the guidelines, promoting consistent, interpretable, and aligned classifications.
    This also allows easy and effective adaptation to different harmfulness criteria across diverse sociocultural contexts with only minor resources and human effort.
    
    Through extensive experiments on benchmark datasets,
    we show that our proposed 
    zero-shot harmful meme detection 
    framework
    effectively narrows the performance gap to state-of-the-art full-shot supervised fine-tuned baselines.
    Notably, with guided CoT prompting, lightweight LLMs (such as Mistral-12B and Qwen2.5-14B) perform competitively on zero-shot inference, matching and even outperforming GPT-4o(-mini) and other low-resource baselines.
We summarize our main contributions as three-fold:
    \begin{itemize}[leftmargin=*,itemsep=0pt,nosep]
        \item
            We propose a harmful meme detection framework that effectively leverages the inherent image understanding abilities, commonsense knowledge and reasoning abilities of lightweight open-source LMMs and LLMs to enable explainable meme classification under low-resource settings.
            By performing zero-shot inference on high-fidelity meme descriptions under guided CoT prompting, our framework can easily adapt to different harmfulness evaluation criteria across diverse sociocultural contexts.
        \item
            We contribute a set of interpretable and targeted guidelines for harmful meme detection across various contexts. We demonstrate that these high-quality yet simple, human-crafted guidelines, designed to capture diverse forms and features of harmfulness, can effectively enhance the zero-shot performance of LLMs in detecting harmful memes.
        \item
            We benchmark the zero-shot harmful meme detection performance of representative lightweight unimodal LLMs.
            We show that our method fully leverages their inherent capabilities, achieving performance comparable and even superior to that of resource-intensive multimodal methods. This highlights an accessible yet effective solution for multimodal tasks in low-resource scenarios.
    \end{itemize}
\begin{figure*}[!t]
  \centering
  \includegraphics[width=0.98\linewidth]{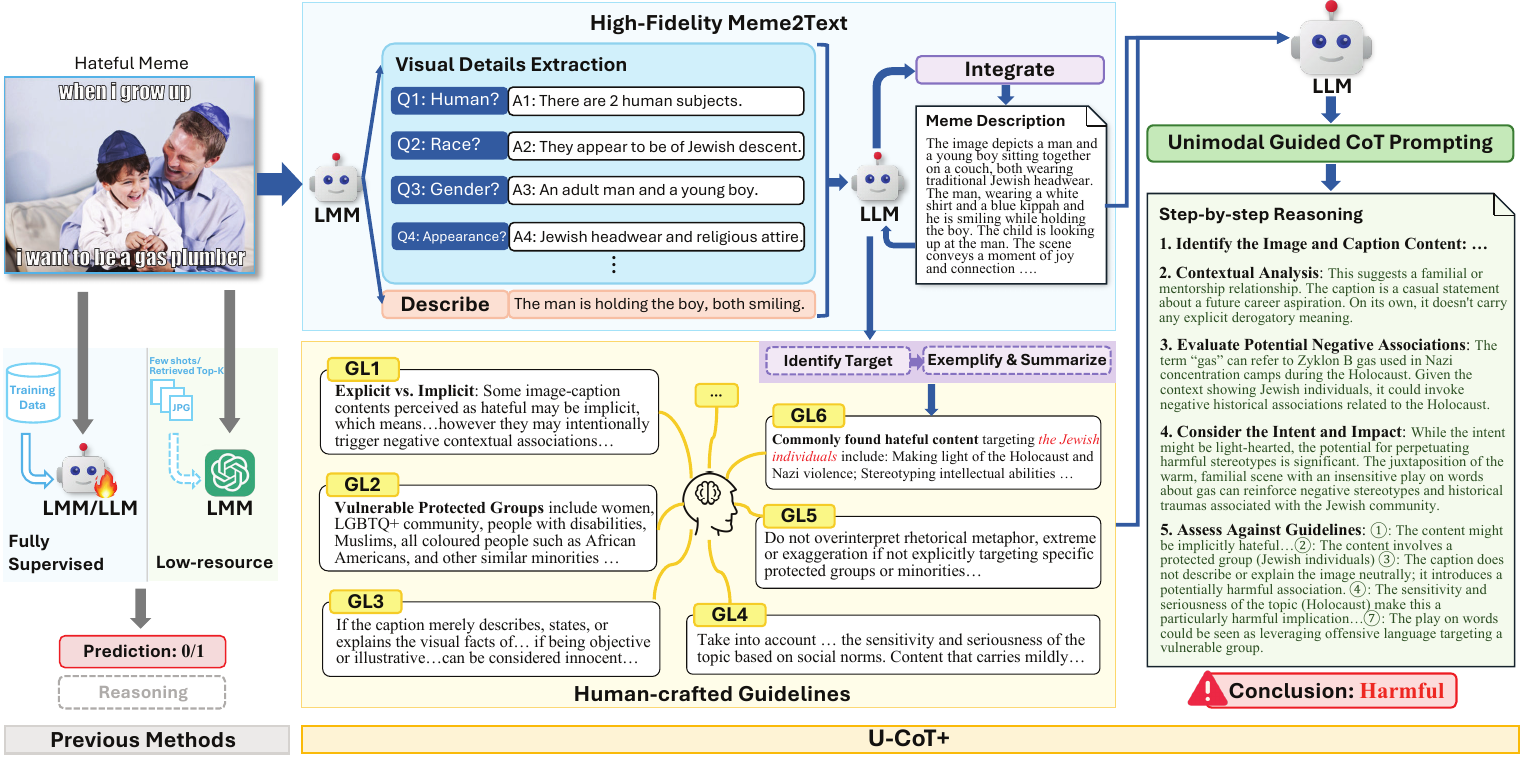}
  \caption{
    Illustration of previous methods and our \textbf{U-CoT+}.
    Previous methods follow either fully supervised or low-resource settings, fine-tuning LMMs/PLMs with labeled data or prompting advanced LMMs (\eg, GPT-4) with/without few-shot examples or retrieval-augmented mechanisms.
    They do not necessarily guarantee predictions that include explicit reasoning.
    U-CoT+ employs a \textbf{High-fidelity Meme2Text} pipeline to convert the multimodal harmful meme detection task into a unimodal, text-only setting, and further enhances LLMs' reasoning through \textbf{Unimodal Guided CoT Prompting}.
    An example output given by Qwen2.5-14B under U-CoT+ is presented in ``Step-by-step Reasoning''.
    \textit{The meme is only for demonstration purposes}.
  }
  \label{fig: main}
\end{figure*}
\section{Related Work}
\paragraph{Harmful Meme Detection.}
    This task seeks to classify online memes as harmful or harmless.
    A major line of works explore contrastive learning models and frameworks (\eg, based on CLIP~\citep{clip}) to train vision language models as multimodal classifiers that fuse cross-modal features and learn unified meme representations in embedding space for image classification
    ~\cite{RGCLnew_mei2025improved, Uncertainty_yang2024uncertainty, HateSieve_su2024hatesieve, ISSUES_burbi2023mapping, tzelepi2025improving, HateClipper_kumar2022hate}.
    Others generate image captions and fine-tune binary text classifiers based on pre-trained language models~\cite{CapAlign_ji2024capalign, LDK_suryawanshi2023multimodal, PromptHate_cao2023prompting}.
    Some injects explainability in detection by prompting advanced LMMs~\cite{Demystifying_hee2025demystifying, ExplainHM_lin2024towards} or LLMs~\cite{M3CoT_kumari2024m3hop} to generate reasoning outputs for supervised training.
    However, existing work on low-resource harmful meme detection remains limited. The use of advanced multimodal models such as LLaVA-34B and GPT-4o as training-free detection agents is explored in~\cite{LowResource_huang2024towards}. Yet, small-scale LMMs struggle with zero-shot detection~\cite{Zeroshot1_rizwan2024zero, Zeroshot2_abdullakutty2024context}. 
    
\textit{LLMs with CoT Reasoning.}
    LLMs extensively pretrained on universal knowledge have demonstrated remarkable zero-shot inference abilities over
    manifold reasoning tasks~\cite{kojima2022large,wei2021finetuned,qi2023large,wu2024topmost,pan2024llms,wu2024akew,wu2024survey,wu2024fastopic,wu2024traco,wu2025sailing,wu2025antileak}.
    The Chain-of-Thought (CoT) method~\cite{wei2022chain} aims to improve inference performance by prompting LLMs to explicitly generate intermediate reasoning steps before arriving at a final verdict.
    Recent studies have shown the effectiveness of CoT-enhanced supervised fine-tuning and LMMs (\eg, GPT-4V) for meme-centered tasks~\cite{CoTMetaphor_xu2024exploring,M3CoT_kumari2024m3hop,GoatBench_lin2024goat}. 
    However, whether and how this applies to lightweight unimodal LLMs for harmful meme detection remains largely unexplored.
\section{Methodology}
    The criteria for defining harmful content vary widely with sociocultural contexts and differ across platforms, regions and even over time.
    The use of diverse specialized policies in real-world content moderation highlights the need for context/domain-specific guidelines.
    By nature, harmful meme detection is a linguistic reasoning task that relies on the understanding of:
    \begin{enumerate*}[label=\textbf{(\roman*)}]
        \item \textbf{Nuances in language use}, including metaphors, sarcasm, humor and other rhetorical devices;
        \item \textbf{Commonsense knowledge} such as social norms, moral values, sociopolitical factors and historical, cultural context;
        \item \textbf{Explicit criteria} that define the decision boundary between harmful and benign content.
    \end{enumerate*}
    Unimodal text-only LLMs, pre-trained on large-scale web corpora and further aligned with human preferences,
    inherently possess extensive world knowledge and fundamental reasoning abilities.
    Trained with refined objectives specifically optimized for general-purpose reasoning and instruction following, they demonstrate more reliable performance on language-only tasks than multimodal counterparts of comparable size.
    Their availability at various scales also offers flexible options for low-resource deployment.

    Therefore, we propose \textbf{U-CoT+}, a unimodal guided CoT framework that aims to harness the inherent abilities of lightweight LLMs in zero-shot inference for low-resource harmful meme detection.
    Adopting an empirically grounded, multi-stage LMM+LLM collaborative workflow~\cite{jha2024memeguard,adak2025memesense,agarwal2024mememqa}, our approach effectively converts the default multimodal classification task into a unimodal, text-based linguistic reasoning setup, allowing LLMs to process memes based on textual descriptions and analyze meme harmfulness guided by human-crafted guidelines. The framework comprises two key components: (1) a \textbf{High-Fidelity Meme2Text} pipeline and (2) \textbf{Unimodal Guided CoT Prompting}, as demonstrated in \Cref{fig: main}.
\subsection{High-Fidelity Meme2Text}
    Typical harmful memes use brief, salient visual summaries to attack people's identities.
    Interpreting their punchlines does not require exhaustive, irrelevant visual details, but rests on a clear understanding of what or who is being targeted.
    Commonly found vulnerable identity-related attributes critical for harmfulness detection include \textit{\textbf{\underline{r}ace, \underline{g}ender, \underline{d}isability, physical \underline{a}ppearance}}, and \textit{\textbf{\underline{c}elebrity}},~\etc~
    However, safety-aligned LMMs, due to debiasing considerations in general-purpose applications, tend to omit socially and culturally sensitive visual cues but use ambiguous, identity-neutral references (\eg, ``a person'' or ``some people'') when describing images with human subjects~\cite{CapAlign_ji2024capalign}, which poses challenges for harmful content detection.
    To retain such essential content, we decompose meme description generation into
    individual steps of visual information extraction via atomic question answering (VQA),
    which does not necessarily require highly capable models.
    Therefore, by repeatedly querying a lightweight LMM (such as LLaVA1.6-7B), each time focusing on a specific attribute,
    we can gather these key information without resorting to more resource-intensive LMMs.
    In this way, we effectively decouple meme content recognition from meme classification reasoning.
    This helps reduce task complexity by avoiding immediate reasoning over complicated, entangled visual elements, which is prone to hallucination.

    Specifically, for each meme $M$, LMMs are given a general ``Describe'' prompt $P^D$ such as ``\textit{What is shown in the meme?}'' to generate a basic low-fidelity description:
    \begin{math}
       D_l = \mathbf{LMM}(M, P^D)
    \end{math},
    and then identify the presence of any human subject. For memes showing humans, LMMs are further prompted to provide identify-related information $I_i = \mathbf{LMM}(M, P^I_i)$, where $i$ denotes different attributes.
    With all visual cues ready, we further leverage unimodal LLMs to integrate all LMM outputs into a unified, compact, and coherent high-fidelity description of the meme content:
    \begin{math}
        D_h = \mathbf{LLM}(D_l, I_r, I_g, I_d, I_a, I_c)
    \end{math},
    which serves as input for the subsequent text-based classification.
\subsection{Unimodal Guided CoT Prompting}
\subsubsection{Guideline Crafting}
    The subjective and context-dependent nature of harmful meme definitions makes it unlikely to eliminate human involvement in guideline crafting process. However, harmful meme detection largely relies on widely accessible commonsense knowledge rather than specialized expert insights. Accordingly, we propose an empirical guideline crafting procedure that incurs only minor human effort:
    \begin{enumerate*}[label=\textbf{(\roman*)}]
        \item We examine the available harmful meme definitions and data annotation guidelines in existing datasets
        for the patterns of harmfulness in representative memes;
        \item We summarize the observed patterns and underlying criteria into coherent natural language guidelines consisting of multiple rules ordered in a logically smooth way (\eg, from general to specific, from features to examples, \etc);
        \item We refine guidelines by probing their clarity and effectiveness on sample memes.
    \end{enumerate*}
    This procedure yields the following set of principles for guideline formulation that are reproducible and transferable across different sociocultural contexts:
\begin{itemize}[leftmargin=*,itemsep=0pt, nosep]
    \item \textbf{Implicitness}
    Harmful memes can be implicit~\cite{LowResource_huang2024towards}.   
    This principle requires LLMs to reflect on the underlying implications of a meme to examine whether it is deliberately crafted to evoke negative contextual interpretations (\eg, associations to harmful stereotypes, sensitive controversies in societal, cultural or political practices \etc)~while wrapping harmful messages beneath deceptively benign visuals and neutral language.
    \item \textbf{Tone \& Intent}
    LLMs tend to interpret memes charitably due to alignment with human preferences for safe responses. To mitigate such tendency, this principle explicitly instructs LLMs to avoid assuming a playful, humorous or light-hearted tone and intent, and instead interpret from a neutral perspective.
    \item \textbf{Fine-grained Taxonomy}
    Context-specific harmful memes often have fine-grained taxonomies of \textit{target} or \textit{type}. 
    A harmful meme's \textbf{target} is the entity toward which its harmful message is directed, such as a vulnerable minority, protected group, organization or specific public figure, \etc~
    The \textbf{type} of a harmful meme refers to the sub-category of harmfulness it falls in, such as ``\textit{stereotypes}'' or ``\textit{body shaming}'' \etc,
    in misogynistic memes \cite{MAMI}.
    Based on such taxonomies, by first prompting LLMs to identify a meme's potential target or breaking down into fine-grained classification sub-tasks,
    this principle seeks to narrow the problem and reduce uncertainty, guiding more focused reasoning through the use of target- or type-specific guidelines.
    \item \textbf{Patterns}
    This principle aims to foster analogical reasoning by prompting LLMs with distilled patterns of harmful memes \eg, ``\textit{Perpetuating stereotypes on female's domestic roles}'', ``\textit{Anti-Semitic contents that make light of the Holocaust}''~\etc
    To derive such patterns with further reduced human efforts, we can leverage LLMs' in-context learning abilities to extend and integrate \allowbreak human\allowbreak -provided examples into structured heuristics.
    \item \textbf{Exception}
    When detecting harmful memes, it is crucial not to prioritize high recall at the cost of precision. This principle aims to prevent over-censorship by introducing exception cases in which memes should be considered ``harmless'' despite containing traces of potentially sensitive content. These exceptions may vary across contexts and domains, often reflecting specific evaluation criteria or inherent biases in data annotation.
\end{itemize}
\subsubsection{Zero-shot CoT Prompting}
    Following the widely adopted implementation of chain-of-thought (CoT) reasoning~\cite{wei2022chain},
    we prompt unimodal LLMs in a \textbf{zero-shot} setting using a heuristic prompt $P^{CoT}$ 
    such as ``\textit{Now, let's think step by step}'' or ``\textit{Now, let's apply the guidelines one by one}''.
    This encourages LLMs to produce the final classification decision $CLS$ 
    through detailed step-by-step rationales $R$ 
    by reasoning over the high-fidelity meme description $D_h$ together with our human-crafted harmful meme detection guidelines $GL$.
    Consequently, we have: $CLS, R = \mathbf{LLM_{\text{U-CoT+}}}(D_h, GL, P^{CoT})$.
\section{Experiments}
    We report the Accuracy (\textbf{Acc.}) and macro-F1 (\textbf{F1}) score for performance evaluation in line with recent studies~\cite{RGCLnew_mei2025improved, LowResource_huang2024towards}.
    As our method extracts LLM predictions based on exact token match, probability-based AUC scores are not applicable. We report the corresponding accuracy results for baselines that evaluate using AUC.
    See \Cref{appendix: implementation} for implementation details.
\begin{enumerate}[label={},leftmargin=0pt,itemsep=1pt,topsep=0pt, partopsep=0pt,nosep]
    \item \textbf{\textit{Datasets.}}
    We evaluate our proposed framework on seven widely used harmful meme detection datasets:
    \begin{enumerate*}[label=(\roman*)]
        \item \textbf{FHM}\cite{FHM_kiela2020hateful},
        \item \textbf{HarMeme}\allowbreak\ \cite{HarmC_pramanick-etal-2021-detecting},
        \item \textbf{Harm-P}\cite{HarmP_pramanick2021momenta},
        \item \textbf{MultiOFF}\cite{MultiOFF_suryawanshi-etal-2020-multimodal},
        \item \textbf{MAMI}\cite{MAMI},
        \item \allowbreak \textbf{PrideMM}\allowbreak \cite{PrideMM_shah2024memeclip} and
        \item \textbf{GoatBench}\cite{GoatBench_lin2024goat}.
    \end{enumerate*}
    These datasets and benchmarks cover various types of harmfulness against diverse targets across a wide range of sociocultural contexts.
    \item \textbf{\textit{Baselines.}}
    For \textbf{full-shot SFT} methods, we consider:
    \begin{enumerate*}[label=(\roman*)]
        \item LMM-RGCL\allowbreak \cite{RGCLnew_mei2025improved},
        \item UMR\cite{Uncertainty_yang2024uncertainty},
        \item \allowbreak ExplainHM\cite{ExplainHM_lin2024towards},
        \item Pro-Cap\cite{Procap_cao2023pro},
        \item IntMeme\allowbreak \cite{Demystifying_hee2025demystifying},
        \item CapAlign\cite{CapAlign_ji2024capalign},
        \item M3Hop-CoT\cite{M3CoT_kumari2024m3hop},
        \item MemeCLIP\cite{PrideMM_shah2024memeclip}.
    \end{enumerate*}
    For \textbf{low-resource training-free} methods, we include
    \begin{enumerate*}[label=(\roman*)]
        \item Zero-shot inference with GPT-4o\allowbreak -mini\allowbreak\cite{openai2024gpt4o},
        \item \textsc{LoReHM}\cite{LowResource_huang2024towards},
        \item \textsc{MInD}\cite{liu2025mind},
        \item Mod-HATE\cite{modhate},
        and 
        \item GPT-4V\cite{openai2023gpt4v}.
    \end{enumerate*}
    \item \textbf{\textit{Models.}}
    We experiment with representative state-of-the-art lightweight open-source models, including LLaVA1.6-7B~\cite{llavaNext} and Qwen2\allowbreak VL-7B\cite{wang2024qwen2vlenhancingvisionlanguagemodels} as LMMs,
    and instruction-tuned Qwen2.5-14B and 7B~\cite{qwen2.5}, Mistral-12B~\cite{mistral-12b}, and Llama3.1-8B~\cite{llama3modelcard} as unimodal LLMs.
\end{enumerate}
\begin{table*}[!ht]
  \caption{
        Comparison with SOTA SFT and low-resource baselines. The best SFT results are in \textbf{bold}. The best zero-/few-shot performance is highlighted in \textbf{\underline{bold and underlined}}; second-best is \underline{underlined}. $^{\dag}$:
        RGCL+RKC/FS is the out-of-distribution performance of LMM-RGCL~\cite{RGCLnew_mei2025improved} with retrieval augmentation and few-shot examples, respectively.
    }
  \centering
  \small
  \setlength{\tabcolsep}{4.4pt}
    \begin{tabular}{l|l|cc|cc|cc|cc|cc|cc}
    \toprule[1pt]
    \textbf{Dataset} & \multicolumn{1}{c|}{\multirow{2}[4]{*}{\textbf{Base Model}}} & \multicolumn{2}{c|}{\textbf{FHM}} & \multicolumn{2}{c|}{\textbf{HarMeme}} & \multicolumn{2}{c|}{\textbf{Harm-P}} & \multicolumn{2}{c|}{\textbf{MultiOFF}} & \multicolumn{2}{c|}{\textbf{MAMI}} & \multicolumn{2}{c}{\textbf{PrideMM}} \\
\cmidrule{1-1}    \textbf{Method} &       & \textbf{Acc.} & \textbf{F1} & \textbf{Acc.} & \textbf{F1} & \textbf{Acc.} & \textbf{F1} & \textbf{Acc.} & \textbf{F1} & \textbf{Acc.} & \textbf{F1} & \textbf{Acc.} & \textbf{F1} \\
    \midrule
    \multicolumn{14}{l}{\textit{\textbf{Full-shot Supervised Fine-tuned Classifiers}}} \\
    \midrule
    LMM-RGCL  & Qwen2VL-7B & \textbf{82.10} & --     & \textbf{88.10} & --     & 91.60 & 91.10 & \textbf{71.10} & \textbf{64.80} & 79.90 & --     & \textbf{78.10} & \textbf{78.40} \\
    UMR   & BLIP  & 77.20 & --     & 87.85 & 86.99 & 92.11 & 92.11 & --     & 69.96 & --     & --     & --     & -- \\
    ExplainHM & GPT3.5+FLAN-T5 & 75.60 & 75.39 & 87.00 & 86.41 & 90.73 & 90.72 & --     & --     & --     & --     & --     & -- \\
    Pro-Cap & BLIP-2+RoBERTa & 75.10 & --     & 85.03 & --     & --     & --     & --     & --     & 73.63 & --     & --     & -- \\
    IntMeme & FLAVA+RoBERTa & 71.52 & --     & 82.66 & --     & --     & --     & --     & --     & 72.44 & --     & --     & -- \\
    CapAlign & ChatGPT+FLAN-T5 & --     & --     & 82.20 & 68.63 & \textbf{93.80} & \textbf{93.11} & --     & --     & --     & --     & --     & -- \\
    M3Hop-CoT & CLIP \etc+Mistral-7B & --     & --     & --     & --     & --     & --     & --     & --     & \textbf{80.28} & --     & --     & -- \\
    MemeCLIP & CLIP  & --     & --     & --     & --     & --     & --     & --     & --     & --     & --     & 76.06 & 75.09 \\
    \midrule
    \multicolumn{14}{l}{\textit{\textbf{Low-resource Baselines w/ Retrieval Augmented Classification}}} \\
    \midrule
    \textsc{LoReHM} & LLaVA-34B & 65.60 & 65.59 & 73.73 & 70.86 & --     & --     & --     & --     & 75.40 & 75.28 & --     & -- \\
    \textsc{LoReHM} & GPT-4o & 70.20 & 70.14 & 74.54 & 72.98 & --     & --     & --     & --     & \textbf{\underline{83.00}} & \textbf{\underline{82.98}} & --     & -- \\
    \textsc{MInD} & LLaVA1.5-13B & 60.80 & 60.71 & 68.93 & 65.19 & --     & --     & --     & --     & 68.90 & 68.84 & --     & -- \\
    RGCL+RKC$^{\dag}$ & Qwen2VL-7B & 69.30 & --     & 81.90 & --     & \textbf{\underline{67.30}} & \textbf{\underline{67.80}} & 63.80 & 55.60 & 75.60 & --     & 69.30 & 69.30 \\
    \midrule
    \multicolumn{14}{l}{\textit{\textbf{Low-resource Baselines w/o Retrieval Augmented Classification}}} \\
    \midrule
    Zero-shot &  GPT-4o-mini  & 67.60 & 65.51 & 70.90 & 69.46 & \underline{65.35} & \underline{65.35} & \underline{65.77} & \underline{64.86} & 77.40 & 76.59 & \textbf{\underline{72.39}} & \textbf{\underline{72.28}} \\
    RGCL+FS$^{\dag}$ & Qwen2VL-7B & 63.50 & --     & 68.10 & --     & --     & --     & --     & --     & --     & --     & --     & -- \\
    Mod-HATE & BLIP-2+LLaMA-7B & 57.60 & 53.88 & 71.19 & 69.64 & --     & --     & --     & --     & 69.05 & 68.78 & --     & -- \\
    \midrule
    \multicolumn{14}{l}{\textit{\textbf{Ours: Unimodal Guided Zero-shot CoT}}} \\
    \midrule
    \multicolumn{1}{l|}{\multirow{2}[1]{*}{U-CoT+}} & LMM$_{\text{7B}}$+Qwen2.5-14B & \underline{72.50} & \underline{72.41} & \underline{83.62} & \underline{82.00} & \underline{65.35} & \underline{65.35} & \textbf{\underline{69.13}} & \textbf{\underline{68.37}} & \underline{79.90} & \underline{79.89} & \underline{71.60} & \underline{71.37} \\
          & LMM$_{\text{7B}}$+Mistral-12B & \textbf{\underline{72.90}} & \textbf{\underline{72.87}} & \textbf{\underline{83.90}} & \textbf{\underline{82.41}} & 63.10 & 62.01 & 64.43 & 61.36 & 75.00 & 74.72 & 69.03 & 68.07 \\
    \bottomrule[1pt]
    \end{tabular}%
  \label{tab:main table}%
\end{table*}%
\begin{table}[!t]
  \caption{
    Performance on Goat-Bench.
    \textbf{S.}: Scheme.
    M: M-CoT; U: U-CoT; U+: U-CoT+.
    GPT-4V results are reported in \cite{GoatBench_lin2024goat}.
    Our best results are highlighted in \textbf{bold}.
   }
  \centering
  \small
  \setlength{\tabcolsep}{1.8pt}
    \begin{tabular}{c|l|rr|rr|rr|rr}
    \toprule[1pt]
    \multirow{2}[2]{*}{\textbf{Model}} & \multicolumn{1}{c|}{\multirow{2}[2]{*}{\textbf{S.}}} & \multicolumn{2}{c|}{\textbf{Hateful}} & \multicolumn{2}{c|}{\textbf{Harmful}} & \multicolumn{2}{c|}{\textbf{Misogyny}} & \multicolumn{2}{c}{\textbf{Offensive}} \\
          &       & \multicolumn{1}{c}{\textbf{Acc.}} & \multicolumn{1}{c|}{\textbf{F1}} & \multicolumn{1}{c}{\textbf{Acc.}} & \multicolumn{1}{c|}{\textbf{F1}} & \multicolumn{1}{c}{\textbf{Acc.}} & \multicolumn{1}{c|}{\textbf{F1}} & \multicolumn{1}{c}{\textbf{Acc.}} & \multicolumn{1}{c}{\textbf{F1}} \\
    \midrule
    \textbf{GPT-4V} & M     & 71.90 & 71.37 & 66.90 & 64.97 & 81.60 & 81.57 & 61.37 & 60.99 \\
    \midrule
    \multirow{2}[1]{*}{\textbf{Qwen2.5-14B}} & U     & 70.80 & 69.17 & 62.64 & 60.51 & 77.40 & 77.38 & 60.97 & 59.76 \\
          & U+    & \textbf{71.95} & \textbf{71.05} & \textbf{71.16} & \textbf{70.69} & \textbf{80.10} & \textbf{80.09} & 61.78 & \textbf{60.90} \\
    \midrule
    \multirow{2}[1]{*}{\textbf{Mistral-12B}} & U     & 70.20 & 68.21 & 61.45 & 58.93 & 75.50 & 75.48 & \textbf{63.53} & 60.62 \\
          & U+    & 71.45 & 70.43 & 70.76 & 68.32 & 75.10 & 74.83 & 60.70 & 58.86 \\
    \bottomrule[1pt]
    \end{tabular}%
  \label{tab:GB}%
\end{table}%
 
    \Cref{tab:main table} and \Cref{tab:GB} compare our best zero-shot harmful meme detection performance achieved by the best-performing unimodal LLMs under U-CoT+ to all the state-of-the-art (SOTA) baselines under either full-shot SFT or low-resource settings.
    We denote the traditional LMM-based multimodal classification scheme under zero-shot CoT as \textbf{M-CoT}, where the LMM reasons over the prompt and image.
    \textbf{U-CoT} denotes \textbf{U-CoT+}'s ablation variant, where the LLM performs zero-shot CoT reasoning over textual inputs but without referring any supportive information such as guidelines.
\begin{table*}[!ht]
  \caption{
    Ablation studies.
    The best low-resource results (with open-source models, without ensemble) are in \textbf{bold}, with the second-best \underline{underlined}.
    $\Delta$: Performance gain over 0-shot guideline-free U-CoT.
    $\bar{\Delta}_{\textit{U-CoT+}}$: The average performance gain of U-CoT+ under guideline perturbations.
    $\Delta_{\textit{U-CoT+L}}$: The performance gain of U-CoT+ with \textsc{LoReHM} insights.
    $\bar{\Delta}_{\textit{U-CoT+FS}}$: The average performance gain of U-CoT+ with few-shot (4,6,8 and 10-shot) examples.
   }
  \centering
  \resizebox{0.96\linewidth}{!}{
  \begin{threeparttable}
    \begin{tabular}{c|c|l|rr|rr|rr|rr|rr|rr}
    \toprule[1pt]
    \multirow{2}[1]{*}{\textbf{LLM}} & \multirow{2}[1]{*}{\textbf{LMM}} & \multicolumn{1}{c|}{\multirow{2}[1]{*}{\textbf{Scheme}}} & \multicolumn{2}{c|}{\textbf{FHM}} & \multicolumn{2}{c|}{\textbf{HarMeme}} & \multicolumn{2}{c|}{\textbf{Harm-P}} & \multicolumn{2}{c|}{\textbf{MultiOFF}} & \multicolumn{2}{c|}{\textbf{MAMI}} & \multicolumn{2}{c}{\textbf{PrideMM}} \\
          &       &       & \multicolumn{1}{c}{\textbf{Acc.}} & \multicolumn{1}{c|}{\textbf{F1}} & \multicolumn{1}{c}{\textbf{Acc.}} & \multicolumn{1}{c|}{\textbf{F1}} & \multicolumn{1}{c}{\textbf{Acc.}} & \multicolumn{1}{c|}{\textbf{F1}} & \multicolumn{1}{c}{\textbf{Acc.}} & \multicolumn{1}{c|}{\textbf{F1}} & \multicolumn{1}{c}{\textbf{Acc.}} & \multicolumn{1}{c|}{\textbf{F1}} & \multicolumn{1}{c}{\textbf{Acc.}} & \multicolumn{1}{c}{\textbf{F1}} \\
    \midrule
    \multirow{3}[4]{*}{N/A} & GPT-4o-mini & \multicolumn{1}{c|}{--} & 67.60 & 65.51 & 70.90 & 69.46 & 65.35 & 65.35 & 65.77 & 64.86 & 77.40 & 76.59 & 72.39 & 72.28 \\
\cmidrule{2-15}          & Qwen2VL-7B & \multirow{2}[2]{*}{M-CoT} & 64.20 & 62.68 & 67.51 & 60.29 & 56.34 & 53.05 & 71.14 & 64.61 & 68.10 & 66.03 & 68.44 & 68.43 \\
          & LLaVA1.6-7B &       & 60.40 & 57.85 & 66.38 & 61.05 & 56.34 & 53.62 & 59.73 & 57.38 & 67.80 & 67.46 & 60.16 & 59.99 \\
    \midrule
    \multirow{7}[4]{*}{\textbf{Qwen2.5-14B}} & \multirow{5}[2]{*}{7B LMM} & U-CoT & \textcolor[rgb]{0,0.5,1}{70.10} & \textcolor[rgb]{0,0.5,1}{70.02} & 59.60 & 51.46 & \textcolor[rgb]{0,0.5,1}{60.00} & \textcolor[rgb]{0,0.5,1}{59.91} & 64.43 & 62.93 & \underline{\textcolor[rgb]{0,0.5,1}{76.40}} & \underline{\textcolor[rgb]{0,0.5,1}{76.38}} & 66.47 & 66.47 \\
          &       & U-CoT+ & \underline{72.50} & \underline{72.41} & \underline{83.62} & \underline{82.00} & \textbf{65.35} & \textbf{65.35} & \textbf{69.13} & \textbf{68.37} & \textbf{79.90} & \textbf{79.89} & \textbf{71.60} & \textbf{71.37} \\
          &       & \multicolumn{1}{r|}{$\bar{\Delta}_{\textit{U-CoT+}}$} & 1.78  & 1.75  & 22.11 & 28.58 & 4.22  & 4.23  & 0.50  & 0.98  & 1.96  & 1.93  & 3.70  & 3.60 \\
          &       & \multicolumn{1}{r|}{$\Delta_{\textit{U-CoT+L}}$} & 0.40  & 0.47  & 7.63  & 14.44 & \multicolumn{1}{c}{--} & \multicolumn{1}{c|}{--} & \multicolumn{1}{c}{--} & \multicolumn{1}{c|}{--} & 0.30  & 0.32  & \multicolumn{1}{c}{--} & \multicolumn{1}{c}{--} \\
          &       & \multicolumn{1}{r|}{$\bar{\Delta}_{\textit{U-CoT+FS}}$} & -0.70 & -0.62 & 4.81  & 1.78  & 1.20  & 1.21  & -0.67 & -1.77 & 1.25  & 1.25  & 0.44  & 0.26 \\
\cmidrule{2-15}          & 7B LMMs & Ensemble & 72.40 & 72.23 & 82.77  & 82.13  & 65.07  & 65.05  & 62.42 & 62.21 & 79.90  & 79.85  & \textcolor[rgb]{1,0,0}{$\uparrow$}73.37  & \textcolor[rgb]{1,0,0}{$\uparrow$}73.36 \\
          & GPT-4o-mini & U-CoT+ & 73.00 & 73.00 & 83.90 & 83.05 & 62.25 & 62.25 & 60.40 & 60.22 & 79.10 & 78.78 & 68.44 & 68.39 \\
    \midrule
    \multirow{7}[4]{*}{\textbf{Mistral-12B}} & \multirow{5}[2]{*}{7B LMM} & U-CoT & \textcolor[rgb]{0,0.5,1}{67.20} & \textcolor[rgb]{0,0.5,1}{66.70} & 65.54 & \textcolor[rgb]{0,0.5,1}{63.48} & \textcolor[rgb]{0,0.5,1}{58.87} & \textcolor[rgb]{0,0.5,1}{58.32} & 61.74 & 56.89 & \textcolor[rgb]{0,0.5,1}{73.00} & \textcolor[rgb]{0,0.5,1}{72.99} & 62.72 & 62.16 \\
          &       & U-CoT+ & \textbf{72.90} & \textbf{72.87} & \textbf{83.90} & \textbf{82.41} & \underline{63.10} & \underline{62.01} & 64.43 & 61.36 & 74.90 & 74.81 & \underline{69.03} & \underline{68.07} \\
          &       & \multicolumn{1}{r|}{$\bar{\Delta}_{\textit{U-CoT+}}$} & 4.22  & 4.70  & 14.76 & 14.94 & 4.51  & 4.04  & 2.69  & 4.43  & 2.20  & 2.09  & 6.46  & 6.49 \\
          &       & \multicolumn{1}{r|}{$\Delta_{\textit{U-CoT+L}}$} & -1.40 & -2.92 & 0.00  & -4.33 & \multicolumn{1}{c}{--} & \multicolumn{1}{c|}{--} & \multicolumn{1}{c}{--} & \multicolumn{1}{c|}{--} & -5.50 & -7.75 & \multicolumn{1}{c}{--} & \multicolumn{1}{c}{--} \\
          &       & \multicolumn{1}{r|}{$\bar{\Delta}_{\textit{U-CoT+FS}}$} & -0.73 & -0.96 & 1.20  & -4.78 & 1.77  & -1.74 & -0.67 & -5.70 & 1.15  & 0.77  & 0.79  & -1.41 \\
\cmidrule{2-15}          & 7B LMMs & Ensemble & \textcolor[rgb]{1,0,0}{$\uparrow$}73.50 & \textcolor[rgb]{1,0,0}{$\uparrow$}73.43 & 83.90  & 82.86 & 63.10  & 62.16 & 63.09 & 61.53 & \textcolor[rgb]{1,0,0}{$\uparrow$}76.00  & \textcolor[rgb]{1,0,0}{$\uparrow$}75.61  & \textcolor[rgb]{1,0,0}{$\uparrow$}70.02  & \textcolor[rgb]{1,0,0}{$\uparrow$}70.02 \\          
          & GPT-4o-mini & U-CoT+ & 71.10  & 71.09 & 81.92 & 80.61 & 62.82 & 61.27 & 67.79 & 66.12 & 75.20  & 74.84 & 66.27 & 66.21 \\
    \midrule
    \multirow{3}[0]{*}{\textbf{Llama3.1-8B}} & \multirow{2}[0]{*}{7B LMM} & \multicolumn{1}{l|}{U-CoT} & 63.50 & 61.44 & 62.43 & 50.64 & \textcolor[rgb]{0,0.5,1}{62.25} & \textcolor[rgb]{0,0.5,1}{58.78} & 63.76 & 57.55 & \textcolor[rgb]{0,0.5,1}{68.70} & \textcolor[rgb]{0,0.5,1}{68.10} & 60.55 & 58.49 \\
          &       & \multicolumn{1}{l|}{U-CoT+} & \textcolor[rgb]{0,0.5,0}{68.00} & \textcolor[rgb]{0,0.5,0}{66.98} & \textcolor[rgb]{0,0.5,0}{71.19} & \textcolor[rgb]{0,0.5,0}{61.44} & \textcolor[rgb]{0,0.5,0}{61.13} & \textcolor[rgb]{0,0.5,0}{56.21} & 68.46 & 62.35 & \textcolor[rgb]{0,0.5,0}{73.50} & \textcolor[rgb]{0,0.5,0}{73.50} & 64.50 & 63.26 \\
\cmidrule{2-15}      & 7B LMMs & \multicolumn{1}{l|}{Ensemble} & \textcolor[rgb]{1,0,0}{$\uparrow$}69.50 & \textcolor[rgb]{1,0,0}{$\uparrow$}69.00 & \textcolor[rgb]{1,0,0}{$\uparrow$}75.14 & \textcolor[rgb]{1,0,0}{$\uparrow$}69.34 & 60.56 & 56.29 & 67.79 & 61.79 & \textcolor[rgb]{1,0,0}{$\uparrow$}74.00 & \textcolor[rgb]{1,0,0}{$\uparrow$}74.00 & \textcolor[rgb]{1,0,0}{$\uparrow$}67.85 & \textcolor[rgb]{1,0,0}{$\uparrow$}67.58 \\
    \midrule
    \multirow{3}[1]{*}{\textbf{Qwen2.5-7B}} & \multirow{2}[0]{*}{7B LMM} & \multicolumn{1}{l|}{U-CoT} & \textcolor[rgb]{0,0.5,1}{66.70} & \textcolor[rgb]{0,0.5,1}{66.45} & 59.89 & 59.82 & \textcolor[rgb]{0,0.5,1}{58.59} & \textcolor[rgb]{0,0.5,1}{57.46} & 63.76 & 63.72 & \textcolor[rgb]{0,0.5,1}{70.60} & \textcolor[rgb]{0,0.5,1}{70.59} & 63.51 & 62.38 \\
          &       & \multicolumn{1}{l|}{U-CoT+} & \textcolor[rgb]{0,0.5,0}{67.50} & \textcolor[rgb]{0,0.5,0}{67.50} & \textcolor[rgb]{0,0.5,0}{77.12} & \textcolor[rgb]{0,0.5,0}{75.26} & \textcolor[rgb]{0,0.5,0}{62.54} & \textcolor[rgb]{0,0.5,0}{60.20} & \underline{67.11} & \underline{\textcolor[rgb]{0,0.5,0}{66.63}} & \textcolor[rgb]{0,0.5,0}{74.40} & \textcolor[rgb]{0,0.5,0}{74.40} & 67.26 & 66.90 \\
\cmidrule{2-15}      & 7B LMMs & Ensemble & 67.00 & 66.98 & 70.34 & 70.23 & \textcolor[rgb]{1,0,0}{$\uparrow$}62.82 & \textcolor[rgb]{1,0,0}{$\uparrow$}61.76 & 63.09 & 63.09 & \textcolor[rgb]{1,0,0}{$\uparrow$}75.80 & \textcolor[rgb]{1,0,0}{$\uparrow$}75.78 & 67.06 & 67.03 \\
    \bottomrule[1pt]
    \end{tabular}%
    \begin{tablenotes}
      \footnotesize
      \item Note: U-CoT results outperforming M-CoT are \textcolor[rgb]{0,0.5,1}{blue}.
      U-CoT+ results based on $\leq$8B LLMs that outperform M-CoT are \textcolor[rgb]{0,0.5,0}{green}.
      \textcolor[rgb]{1,0,0}{$\uparrow$}: Ensemble results further improving on U-CoT+. 
    \end{tablenotes}
    \end{threeparttable}
    }
  \label{tab:Ablation}%
\end{table*}%
\subsection{Comparing U-CoT+ to Baselines}
\textbf{U-CoT+ vs. Supervised Fine-tuned Baselines}
    As shown in \Cref{tab:main table}, performance gaps remain between the state-of-the-art fully supervised fine-tuned baselines and lightweight LLMs under our proposed unimodal guided zero-shot CoT prompting, but they have narrowed to a generally acceptable and promising range, especially given our focus on low-resource settings.
    Specifically, LLMs' zero-shot performance has become nearly comparable to that of full-shot baselines on FHM, HarMeme, MAMI and PrideMM.
    On FHM, Mistral-12B outperforms IntMeme and comes within 3 points of the performance of ProCap and ExplainHM.
    Notably, its performance on HarMeme
    is highly comparable to that of most SFT baselines, surpassing IntMeme and CapAlign, even with better macro-F1.
    Moreover, both LLMs perform well on MAMI, with Qwen2.5-14B matching the performance of LMM-RGCL. 
    Its best zero-shot result on the challenging PrideMM also falls within 7 points to that of the two SFT baselines.
    On the U.S. political meme dataset Harm-P, SFT methods consistently achieve high accuracy over 90\%, whereas zero-shot inference, even with GPT-4o-mini, struggles to exceed 70\%, which reflects the complex, highly subjective, heterogeneous, and target-dependent nature of defining harmfulness in Harm-P.
    In contrast, SFT methods perform less strongly on MultiOFF, despite its contextual and stylistic similarity to Harm-P, making Qwen2.5-14B's zero-shot performance comparable.
    SFT's decline on MultiOFF suggests a drawback of reduced training data (455 vs. 2.9K in Harm-P).
    This observation verifies SFT methods' heavy reliance on large-scale training data, revealing their inherent limitations under low-resource conditions.

\textbf{U-CoT+ vs. Low-Resource Baselines}
    Low-resource baselines with retrieval-based augmentation such as \textsc{LoReHM}, \textsc{MInD}, though operating under a zero-shot setting, further benefit from a majority voting of ground-truths of the top-K most similar memes retrieved from the embedding space.
    In contrast, U-CoT+ operates under a more constrained low-resource setting, without additional access to auxiliary ground-truth signals from neighbour memes as reference for decision makings.
    Despite the absence of retrieval augmentation, we show that lightweight LLMs can unlock their full potential for zero-shot harmful meme detection.
    As shown in \Cref{tab:main table}, with the support of our human-crafted guidelines, both Qwen2.5-14B and Mistral-12B can largely outperform \textsc{MInD}, and can outperform the other two retrieval-augmented baselines on FHM and HarMeme, even surpassing \textsc{LoReHM}'s performance based on GPT-4o.
    Qwen2.5-14B can further outperform LMM-RGCL+RKC on MultiOFF, MAMI and PrideMM.
    Notably, the diminished out-of-distribution performance of LMM-RGCL, even when enhanced by the retrieval mechanism (RKC), further demonstrates the limited generalizability of SFT methods.  
    Furthermore, U-CoT+ can also significantly outperform LMM-RGCL+FS and Mod-HATE, both of which are baselines without retrieval augmentation but are evaluated under few-shot settings.
    Besides, lightweight LLMs under U-CoT+ consistently outperform or match GPT-4o-mini's zero-shot performance based on multimodal inference across five of six datasets, with slightly weaker results only observed on PrideMM.
    \Cref{tab:GB} compares the zero-shot performance of our method to that of GPT-4V on Goat-Bench~\cite{GoatBench_lin2024goat}.
    As can be seen, U-CoT+ enables lightweight unimodal LLMs to achieve performance highly comparable to, and in some cases even surpassing that of the advanced multimodal GPT-4V across all four sub-tasks.

\textbf{In summary}, results validate U-CoT+'s advantages:
    \begin{enumerate*}[label=\textbf{(\roman*)}]
        \item 
            It effectively leverages lightweight LMMs and LLMs,
            and achieves highly comparable zero-shot performance relative to more resource-intensive baselines, thereby demonstrating strong resource efficiency.
        \item
            Its underlying pipeline and principles support empirical, context-specific adaptation to diverse detection criteria with only minor human efforts, offering flexibility.
        \item
            Its consistently top-ranked performance among low-resource methods across diverse datasets underscores the scalability of our guideline crafting process, which generalizes well across varied sociocultural contexts.
    \end{enumerate*}
\begin{figure}[!t]
  \centering
  \includegraphics[width=0.6\linewidth]{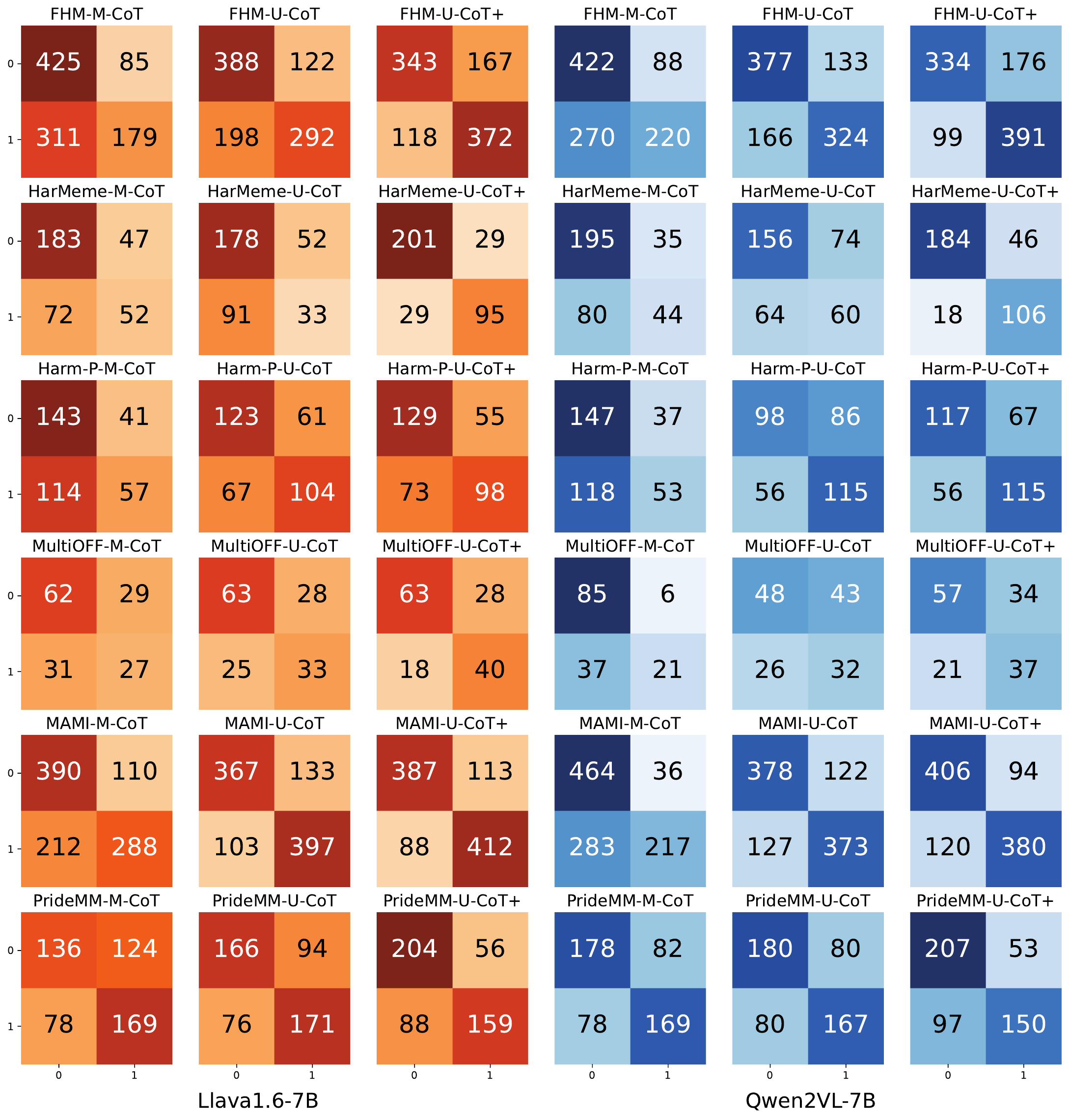}
  \caption{
    Confusion matrices of Qwen2.5-14B based on meme descriptions sourced from different 7B LMMs. 
  }
  \label{fig: confusion matrix}
\end{figure}
\vspace{-10pt}
\subsection{Ablation Study}
    \Cref{tab:Ablation} presents the main results of ablation studies that compare different prompting schemes and validate the effectiveness of key framework components.
    \Cref{fig: confusion matrix} presents the detailed confusion matrices by Qwen2.5-14B.

\textbf{Unimodal vs. Multimodal.}
    We observe consistently improved zero-shot performance on FHM, Harm-P, and MAMI merely by converting the default multimodal setting to unimodal, text-based inference. This highlights the effectiveness of modality conversion, which enables certain meme messages to be more explicitly revealed when conveyed in text.
    \Cref{fig: confusion matrix} shows that both 7B LMMs exhibit an inherent bias toward the negative class, frequently overclassifying memes as ``harmless'' in most datasets.
    This evidences their limitation in detecting harmfulness based on visual inputs.
    Reasoning with more capable unimodal LLMs can, in most cases, help balance classification with more memes revealed to be ``harmful''.

\textbf{Effectiveness \& Robustness.}
    Comparing U-CoT+ against U-CoT results demonstrates the overall effectiveness of our human-crafted guidelines.
    Notably, U-CoT+ enables LLMs of comparable size to 7B LMMs to outperform M-CoT results across four datasets.
    We further apply different perturbation strategies \eg, \textit{Rephrasing}, \textit{Shuffling}, and \textit{Modifying} to our guidelines and compute the resulting average performance gain of U-CoT+ over U-CoT.
    As shown by $\bar{\Delta}_{\textit{U-CoT+}}$ in \Cref{tab:Ablation},
    despite the potential randomness, semantic nuances and logic flow disruption introduced,
    our guidelines remain conceptually constructive, consistently improving performance across most datasets.
    This also evidences U-CoT+'s general robustness to guideline perturbations.
    
\textbf{Human-crafted vs. GPT-generated.}
    We compare the effectiveness of our human-crafted guidelines with GPT-4o-generated harmful meme detection insights from \textsc{LoReHM}~\cite{LowResource_huang2024towards}.
    As show in \Cref{tab:Ablation} ($\bar{\Delta}_{\textit{U-CoT+}}$ vs. $\Delta_{\textit{U-CoT+L}}$), prompting LLMs with human-crafted guidelines in U-CoT+ yields greater improvements over zero-shot U-CoT than using GPT-generated insights,
    which may even lead to performance drops when applied to Mistral-12B.
    These GPT-4o-summarized insights alone, which tend to be overly general and verbose, lacking explicit context-specific relevance and rule-level diversity, prove insufficient for effectively guiding lightweight unimodal LLMs in harmful meme detection.
    This further underscores the usability and effectiveness of our guideline crafting principles in providing targeted, interpretable guidance that is easy to follow.
    
\textbf{Guidelines vs. Few shots.}
    We evaluate four few-shot (FS) settings (4, 6, 8 and 10-shot) with class-balanced meme descriptions randomly sampled from the holdout pool
    (see \Cref{appendix:guideline vs. few shots} for details of demonstration example construction.),
    and compare their average performance gain over zero-shot U-CoT ($\bar{\Delta}_{\textit{U-CoT+FS}}$) to $\bar{\Delta}_{\textit{U-CoT+}}$ in \Cref{tab:Ablation}.
    Results show that, despite occasional improvements, providing LLMs with meme descriptions as few-shot demonstration examples is generally less effective than guideline-based prompting, and may even result in deteriorated performance compared to that of zero-shot inferences, suggesting that LLMs' in-context learning ability does not translate well to this task, which aligns with previous findings for LMMs~\cite{RGCLnew_mei2025improved,LowResource_huang2024towards}.

\textbf{7B LMMs vs. GPT-4o-mini for Meme2Text.}
    We replace meme descriptions produced by our proposed pipelines with GPT-4o-mini-generated descriptions, which generically provide more accurate visual information.
    Results for U-CoT+ in \Cref{tab:Ablation} show that reasoning over meme descriptions generated by the more capable GPT-4o-mini does not consistently outperform those produced by less advanced lighter-weight LMMs,
    suggesting that the performance bottleneck is not primarily visual.
    This further indicates that our proposed meme2text pipeline, despite relying on low-resource models, is still effective in capturing visual information essential for harmfulness assessment.

\textbf{Single vs. Ensemble.}
    We compare single-LMM U-CoT+ with ensemble strategies that aggregate predictions from other scheme variants (\ie, M-CoT and U-CoT), as well as from U-CoT+ based on different LMMs (See \Cref{tab:ensemble_strategies}).
    Results in \Cref{tab:Ablation} show that
    even without relying only on a single best-performing 7B LMM,
    we can largely preserve performance through appropriate ensembles of predictions, which may even yield further improved results on MAMI and PrideMM.
    In practice, ensemble strategies help mitigate uncertainty in LLMs’ sensitivity to the choice of 7B LMMs for visual information extraction, while also alleviating potential over-censorship introduced by strict guideline-based assessment. 
\begin{figure}[!t]
    \centering
    \includegraphics[width=0.9\linewidth]{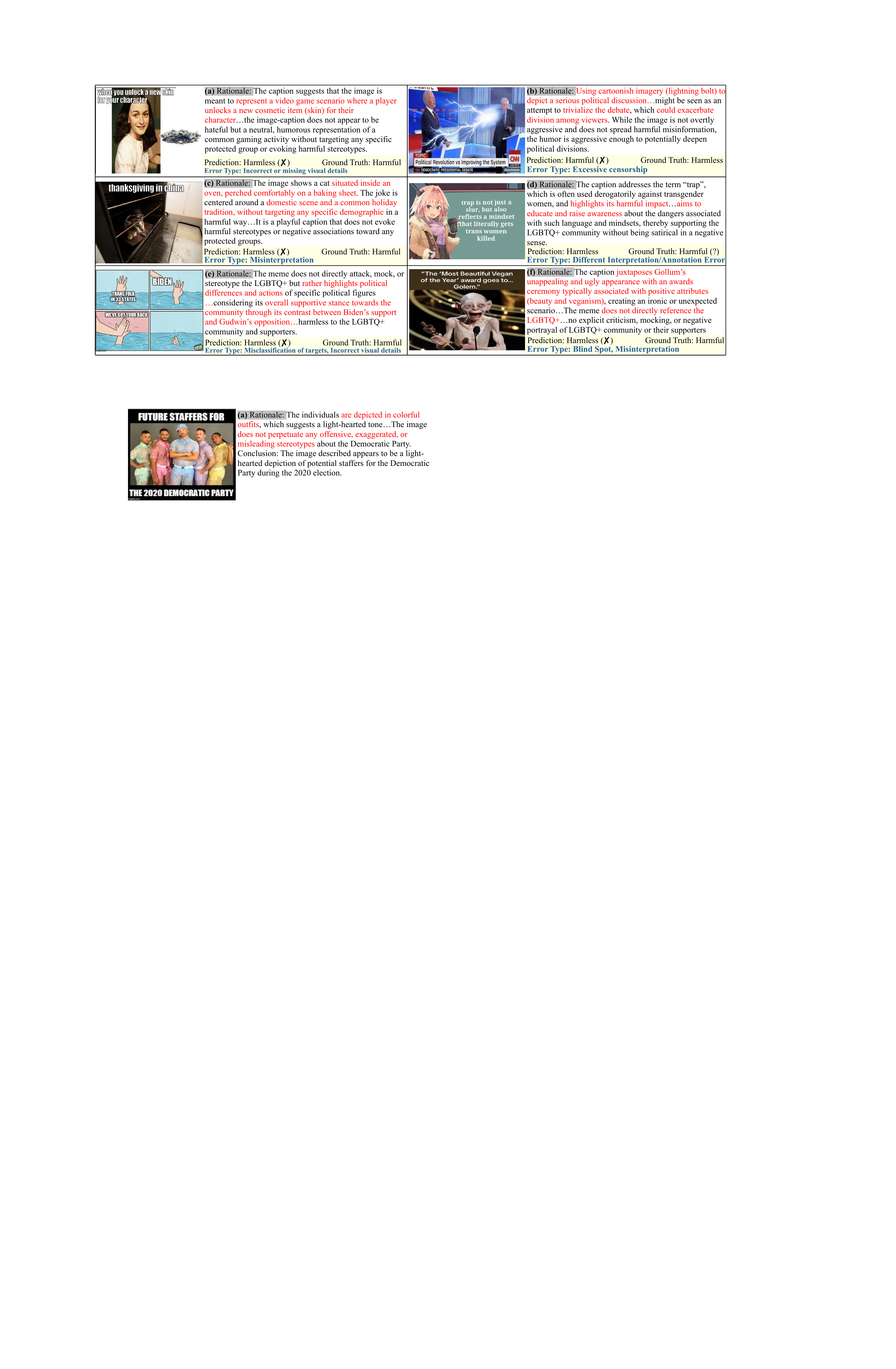}
    \caption{Comparing LLM rationales before and after applying context-specific guidelines.
    \colorbox{mygray}{Incorrect}.
    \colorbox{mygreen}{Correct}.
    }
    \label{fig:before_after_GL}
\end{figure}
\begin{figure*}
    \centering
    \includegraphics[width=0.82\linewidth]{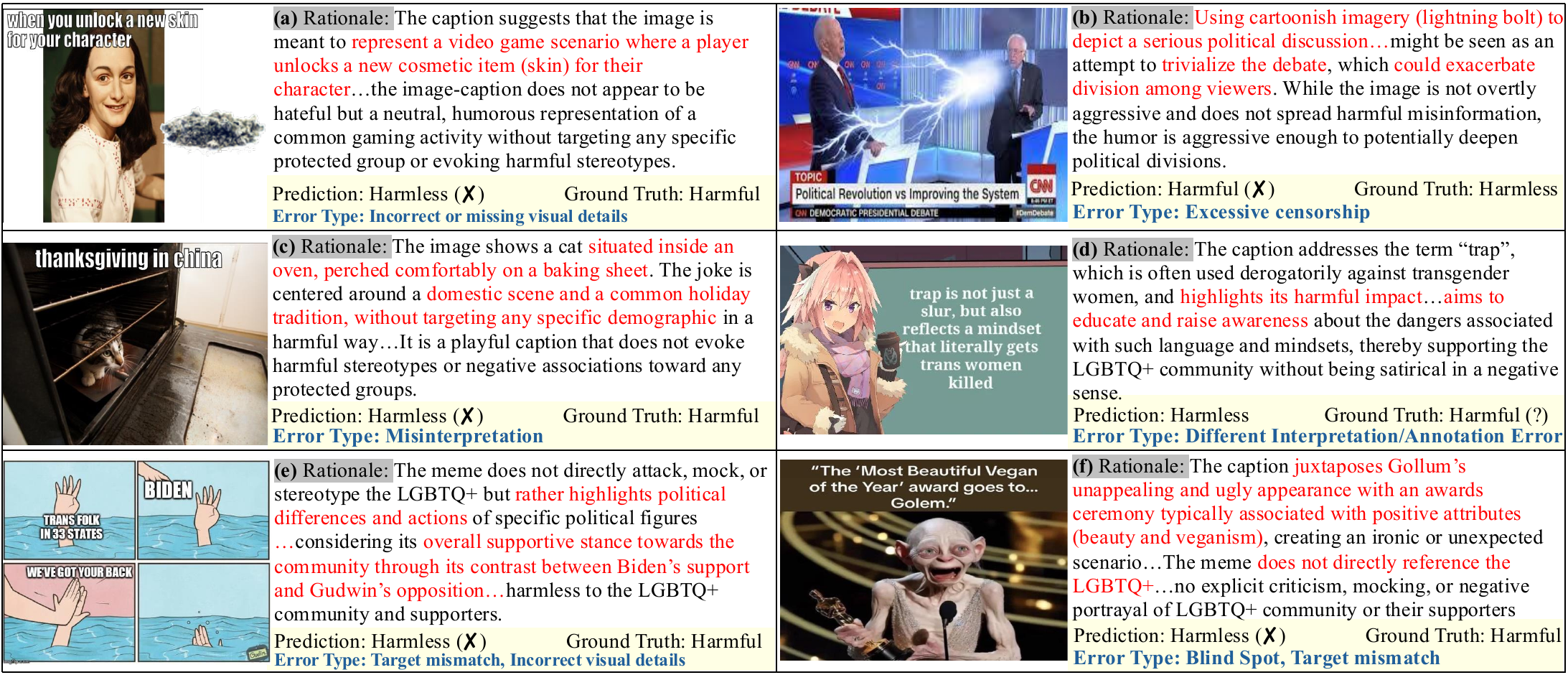}
    \caption{Examples of incorrectly classified memes and their corresponding error type.}
    \label{fig:error_examples}
\end{figure*}
\section{Case Study}
    Although grounded in a low-resource setup, U-CoT+ fully leverages the inherent knowledge and reasoning capabilities of lightweight LLMs for explainable harmful meme detection by prompting them to generate explicit step-by-step rationales leading to the final predictions.
    Comparing LLMs' reasoning trajectories enables a closer examination of how our human-crafted guidelines influence their ``mindset'', as illustrated in \Cref{fig:before_after_GL}.
    First, the guidelines provide LLMs with more aligned and context-specific guidance that enhances their associations, particularly for those negative interpretations with implicit harmful cues.
    For example, in \Cref{fig:before_after_GL}(a), when no guideline is provided, despite correctly recognizing the visual content, the LLM misinterprets the mention of ``plane'' as ``owning a plane'', linking it to a neutral stereotype that assumes Middle Eastern individuals as wealthy. 
    In contrast, when guidelines are presented, the LLM correctly identifies the sensitivity of the ``plane'' reference and associates it with a more serious implication of terrorism by pointing out its possible allusion to the 9/11 attacks, which is crucial for determining the meme as harmful.
    Second, the guidelines clarify the boundary between black and white by defining detailed checklists, thereby reducing uncertainty in the ``gray zone''. 
    For the meme shown in \Cref{fig:before_after_GL}(b), when guidelines are absent, the LLM reasonably acknowledges the meme's potential offensiveness but ultimately downplays it as common political satire. 
    With the guidelines in place, however, the LLM becomes more confident in affirming the meme as offensive based on matches with multiple criteria, such as misinformation, stereotyping, and personal attacks \etc~
    As such, U-CoT+ enables human content moderators to intuitively formulate and apply detection guidelines, and to further refine them or verify models' verdicts based on reasoning outputs, thereby supporting more consistent and accountable decisions.
\section{Error Analysis}
    U-CoT+'s transparency in decision makings enables fine-grained error analysis that reveals the inherent behaviors of lightweight LLMs (and LMMs) in harmful meme detection.
    We summarize common causes of misclassification by closely examining LLMs' rationales.
    \Cref{fig:error_examples} presents typical misclassified examples and \Cref{fig:error_dist} shows the quantitative distribution of different error types:
    \begin{enumerate*}[label=\textbf{(\roman*)}]
        \item
            \textbf{Incorrect or missing visual details}:
            The image understanding and text recognition capabilities of 7B LMMs remain limited. For instance, the LMM fails to recognize Anne Frank in \Cref{fig:error_examples}(a), thereby missing the connection between the visual element ``ash'' and the insensitive caption. Moreover, LMMs still struggle to provide accurate details for visually noisy, stylistically abstract, blurred, or text-heavy memes, such as edited screenshots of chats or tweets.
        \item
            \textbf{Excessive censorship}: 
            LLMs tend to overinterpret or overstate potential negative effects when they apply the guidelines too rigidly or adopt an overly strict perspective, such as interpreting sarcasm or parody in political debates shown in \Cref{fig:error_examples}(b) as trivializing serious issues.
        \item
            \textbf{Misinterpretation}:
            LLMs may misinterpret a meme's tone or intent
            when the meme is subtle, implicit, ambiguous, or poorly framed. Discrepancies between LLMs' and human interpretations of the guidelines can also cause LLMs to overlook or inconsistently apply them.
        \item
            \textbf{Different interpretations}:
            LLMs may arrive at reasonable interpretations that diverge from human annotations, largely due to differences in perceived levels of harm.
            In such cases, LLM judgments are not necessarily incorrect. For instance, LLMs may classify some annotated harmful memes as harmless by interpreting them as mild ridicule or satire that is common and generally acceptable in political discourse. These misclassifications may instead reflect inconsistent or erroneous annotations.
        \item
            \textbf{Target mismatch}:
            The inferred target subject of a meme determines which subset of guidelines is applied. 
            Discrepancy in target identification can lead to guideline mismatches and subsequent misclassifications.
            \Cref{fig:error_examples}(e) is labeled as harmful toward President Biden, but the LLM instead interprets the target as the LGBTQ+ community and thus predicts the meme as harmless.
        \item
            \textbf{Blind Spot}:
            LLMs adhere to guideline-bounded reasoning and do not actively extend beyond it to account for long-tailed harmfulness types outside the main context.
            In \Cref{fig:error_examples}(f),
            the LLM fails to generalize to the vegan group under the PrideMM guidelines, which primarily focus on LGBTQ+ issues.
    \end{enumerate*}
    These errors highlight limitations in both models and datasets, motivating future improvements:
    \begin{enumerate*}[label=\textbf{(\roman*)}]
        \item Enabling general LLMs to rapidly learn and adapt to evolving popular culture and communication practices; and
        \item Developing multi-level, fine-grained, caption- and explanation-aware annotation frameworks for high-quality harmful meme detection benchmarks.
    \end{enumerate*}
\section{Conclusion}
    In this paper, we introduce U-CoT+, a novel framework for low-resource harmful meme detection. 
    With a high-fidelity meme2text pipeline, it decouples meme content recognition from harmfulness analysis and converts the default multimodal task into a unimodal, text-based setting, thus enabling detection based on LLMs' inherent capabilities.
    By incorporating targeted, interpretable and easily obtained human-crafted guidelines, U-CoT+ further enhances the zero-shot performance of lightweight LLMs with guided CoT prompting, achieving performance highly competitive with, or even surpassing that of resource-intensive baselines across datasets from diverse sociocultural contexts. Experiment results demonstrate U-CoT+'s effectiveness in offering strong resource efficiency, flexibility, scalability and explainability for harmful meme detection.
\section{Acknowledgments}
This work is supported by AI-X, Interdisciplinary Graduate Programme, and the College of Computing and Data Science of Nanyang Technological University, Singapore.
\bibliographystyle{ACM-Reference-Format}
\balance
\bibliography{sample-base}
\appendix
\begin{table*}[!ht]
  \caption{
  Full experiment results.
  $\Delta$: Performance gain over zero-shot guideline-free U-CoT.
  }
  \centering
  \resizebox{0.7\linewidth}{!}{
  \begin{threeparttable}
    \begin{tabular}{c|c|l|rr|rr|rr|rr|rr|rr}
    \toprule[1pt]
    \multirow{2}[2]{*}{\textbf{LLM}} & \multirow{2}[2]{*}{\textbf{LMM}} & \multirow{2}[2]{*}{\textbf{Scheme}} & \multicolumn{2}{c|}{\textbf{FHM}} & \multicolumn{2}{c|}{\textbf{HarMeme}} & \multicolumn{2}{c|}{\textbf{Harm-P}} & \multicolumn{2}{c|}{\textbf{MultiOFF}} & \multicolumn{2}{c|}{\textbf{MAMI}} & \multicolumn{2}{c}{\textbf{PrideMM}} \\
          &       &       & \textbf{Acc.} & \textbf{F1} & \textbf{Acc.} & \textbf{F1} & \textbf{Acc.} & \textbf{F1} & \textbf{Acc.} & \textbf{F1} & \textbf{Acc.} & \textbf{F1} & \textbf{Acc.} & \textbf{F1} \\
    \midrule
    \multirow{4}[4]{*}{N/A} & GPT-4o-mini & --    & 67.60 & 65.51 & 70.90 & 69.46 & 65.35 & 65.35 & 65.77 & 64.86 & 77.40 & 76.59 & 72.39 & 72.28 \\
\cmidrule{2-15}          & Qwen2VL-7B & \multirow{3}[2]{*}{M-CoT} & 64.20 & 62.68 & 67.51 & 60.29 & 56.34 & 53.05 & 71.14 & 64.61 & 68.10 & 66.03 & 68.44 & 68.43 \\
          & LLaVA1.6-7B &       & 60.40 & 57.85 & 66.38 & 61.05 & 56.34 & 53.62 & 59.73 & 57.38 & 67.80 & 67.46 & 60.16 & 59.99 \\
          & LLaVA1.6-13B &       & 63.60 & 62.91 & 60.17 & 57.97 & 56.90 & 56.45 & 54.36 & 53.61 & 72.90 & 72.90 & 56.02 & 55.26 \\
    \midrule
    \multicolumn{1}{c|}{\multirow{12}[6]{*}{\begin{sideways}\textbf{Qwen2.5-14B}\end{sideways}}} & \multirow{4}[2]{*}{Qwen2VL-7B} & U-CoT & 70.10 & 70.02 & 61.02 & 57.92 & 60.00 & 59.91 & 53.69 & 53.15 & 75.10 & 75.10 & 68.44 & 68.42 \\
          &       & U-CoT+ & 72.50 & 72.41 & 81.92 & 81.00 & 65.35 & 65.35 & 63.09 & 62.41 & 76.10 & 75.74 & 70.41 & 70.04 \\
          &       & $\bar{\Delta}_{\textit{U-CoT+}}$ & 1.78  & 1.75  & 21.11 & 23.32 & 4.22  & 4.23  & 10.24 & 10.16 & 2.03  & 1.86  & 2.27  & 2.07 \\
          &       & $\bar{\Delta}_{\textit{U-CoT+FS}}$ & -0.70 & -0.62 & 6.35  & 6.17  & 1.20  & 1.21  & 5.37  & 4.27  & 2.70  & 2.70  & 1.28  & 1.02 \\
\cmidrule{2-15}          & \multirow{4}[2]{*}{LLaVA1.6-7B} & U-CoT & 68.00 & 67.70 & 59.60 & 51.46 & 63.94 & 63.84 & 64.43 & 62.93 & 76.40 & 76.38 & 66.47 & 66.47 \\
          &       & U-CoT+ & 71.50 & 71.48 & 83.62 & 82.00 & 63.94 & 63.67 & 69.13 & 68.37 & 76.90 & 76.77 & 71.60 & 71.37 \\
          &       & $\bar{\Delta}_{\textit{U-CoT+}}$ & 3.22  & 3.51  & 22.11 & 28.58 & -1.33 & -1.70 & 0.50  & 0.98  & 1.57  & 1.53  & 3.70  & 3.60 \\
          &       & $\bar{\Delta}_{\textit{U-CoT+FS}}$ & -0.28 & -0.05 & 4.81  & 1.78  & -1.69 & -1.66 & -0.67 & -1.77 & 1.25  & 1.25  & 0.44  & 0.26 \\
\cmidrule{2-15}          & \multirow{4}[2]{*}{LLaVA + Qwen} & U-CoT & 69.10 & 68.83 & --    & --    & --    & --    & --    & --    & --    & --    & --    & -- \\
          &       & U-CoT+ & 72.00 & 71.98 & --    & --    & --    & --    & --    & --    & --    & --    & --    & -- \\
          &       & $\bar{\Delta}_{\textit{U-CoT+}}$ & 2.95  & 3.18  & --    & --    & --    & --    & --    & --    & --    & --    & --    & -- \\
          &       & $\bar{\Delta}_{\textit{U-CoT+FS}}$ & -1.13 & -0.90 & --    & --    & --    & --    & --    & --    & --    & --    & --    & -- \\
    \midrule
    \multirow{12}[6]{*}{\begin{sideways}\textbf{Mistral-12B}\end{sideways}} & \multirow{4}[2]{*}{Qwen2VL-7B} & U-CoT & 68.30 & 68.09 & 65.54 & 63.48 & 58.87 & 58.32 & 65.10 & 62.53 & 73.00 & 72.99 & 62.72 & 62.16 \\
          &       & U-CoT+ & 71.20 & 71.19 & 83.90 & 82.41 & 63.10 & 62.01 & 59.06 & 57.33 & 73.70 & 73.36 & 69.03 & 68.07 \\
          &       & $\bar{\Delta}_{\textit{U-CoT+}}$ & 3.98  & 4.18  & 14.76 & 14.94 & 4.51  & 4.04  & -3.02 & -2.74 & 2.28  & 2.16  & 6.46  & 6.49 \\
          &       & $\bar{\Delta}_{\textit{U-CoT+FS}}$ & 0.33  & 0.17  & 1.20  & -4.78 & 1.77  & -1.74 & -0.33 & -3.22 & 1.15  & 0.77  & 0.79  & -1.41 \\
\cmidrule{2-15}          & \multirow{4}[2]{*}{LLaVA1.6-7B} & U-CoT & 65.30 & 64.60 & 62.43 & 54.60 & 58.03 & 56.53 & 61.74 & 56.89 & 76.40 & 76.38 & 62.33 & 61.68 \\
          &       & U-CoT+ & 70.00 & 69.88 & 81.07 & 78.57 & 59.72 & 56.40 & 64.43 & 61.36 & 75.60 & 75.33 & 68.05 & 67.82 \\
          &       & $\bar{\Delta}_{\textit{U-CoT+}}$ & 4.60  & 5.20  & 16.95 & 21.59 & 1.41  & -0.26 & 2.69  & 4.43  & 0.82  & 0.77  & 5.82  & 6.33 \\
          &       & $\bar{\Delta}_{\textit{U-CoT+FS}}$ & -0.30 & -0.52 & 4.24  & -1.23 & -1.20 & -6.24 & -0.67 & -5.70 & -1.88 & -2.14 & 2.17  & 0.65 \\
\cmidrule{2-15}          & \multirow{4}[2]{*}{LLaVA + Qwen} & U-CoT & 67.20 & 66.70 & --    & --    & --    & --    & --    & --    & --    & --    & --    & -- \\
          &       & U-CoT+ & 72.90 & 72.87 & --    & --    & --    & --    & --    & --    & --    & --    & --    & -- \\
          &       & $\bar{\Delta}_{\textit{U-CoT+}}$ & 4.22  & 4.70  & --    & --    & --    & --    & --    & --    & --    & --    & --    & -- \\
          &       & $\bar{\Delta}_{\textit{U-CoT+FS}}$ & -0.73 & -0.96 & --    & --    & --    & --    & --    & --    & --    & --    & --    & -- \\
    \midrule
    \multirow{6}[6]{*}{\begin{sideways}\textbf{Llama3.1-8Bf}\end{sideways}} & \multirow{2}[2]{*}{Qwen2VL-7B} & U-CoT+ & 68.00 & 66.98 & 71.19 & 61.44 & 61.13 & 56.21 & 68.46 & 62.35 & 71.90 & 71.86 & 63.51 & 60.86 \\
          &       & ${\Delta}_{\textit{U-CoT+}}$ & 4.50  & 5.54  & 8.76  & 10.80 & -1.12 & -2.57 & 4.70  & 4.80  & 2.10  & 2.77  & 2.17  & 2.96 \\
\cmidrule{2-15}          & \multirow{2}[2]{*}{LLaVA1.6-7B} & U-CoT+ & 67.00 & 65.51 & 72.03 & 60.88 & 57.75 & 51.09 & 63.76 & 53.86 & 73.50 & 73.50 & 64.50 & 63.26 \\
          &       & ${\Delta}_{\textit{U-CoT+}}$ & 4.00  & 5.15  & 6.78  & 14.22 & -1.97 & -5.31 & 0.67  & 2.02  & 4.80  & 5.40  & 3.95  & 4.77 \\
\cmidrule{2-15}          & \multirow{2}[2]{*}{LLaVA+Qwen} & U-CoT+ & 65.70 & 64.10 & --    & --    & --    & --    & --    & --    & --    & --    & --    & -- \\
          &       & ${\Delta}_{\textit{U-CoT+}}$ & 2.90  & 3.95  & --    & --    & --    & --    & --    & --    & --    & --    & --    & -- \\
    \midrule
    \multirow{6}[6]{*}{\begin{sideways}\textbf{Qwen2.5-7Bf}\end{sideways}} & \multirow{2}[2]{*}{Qwen2VL-7B} & U-CoT+ & 67.50 & 67.50 & 70.34 & 69.97 & 62.54 & 60.20 & 62.42 & 62.13 & 73.60 & 73.39 & 67.26 & 66.90 \\
          &       & ${\Delta}_{\textit{U-CoT+}}$ & 0.80  & 1.05  & 21.75 & 22.90 & 3.95  & 2.74  & 5.37  & 5.32  & 1.80  & 1.59  & 3.75  & 4.52 \\
\cmidrule{2-15}          & \multirow{2}[2]{*}{LLaVA1.6-7B} & U-CoT+ & 65.70 & 65.53 & 77.12 & 75.26 & 59.15 & 54.99 & 67.11 & 66.63 & 74.40 & 74.40 & 64.30 & 64.00 \\
          &       & ${\Delta}_{\textit{U-CoT+}}$ & 3.00  & 3.59  & 17.23 & 15.44 & 1.12  & -2.81 & 3.35  & 2.91  & 3.80  & 3.81  & 0.99  & 2.02 \\
\cmidrule{2-15}          & \multirow{2}[2]{*}{LLaVA+Qwen} & U-CoT+ & 66.60 & 66.55 & --    & --    & --    & --    & --    & --    & --    & --    & --    & -- \\
          &       & ${\Delta}_{\textit{U-CoT+}}$ & 2.80  & 3.32  & --    & --    & --    & --    & --    & --    & --    & --    & --    & -- \\
    \bottomrule[1pt]
    \end{tabular}%
    \begin{tablenotes}
      \footnotesize
      \item $\bar{\Delta}_{\textit{U-CoT+}}$: The average performance gain of U-CoT+ under guideline perturbations.
      $\bar{\Delta}_{\textit{U-CoT+FS}}$: The average performance gain of U-CoT+ with few-shot examples.
    \end{tablenotes}
    \end{threeparttable}
    }
  \label{tab:full_results}%
\end{table*}%
\section{\MakeUppercase{Ethics Statement}}
    This work aims to fully harness the inherent capabilities of state-of-the-art LLMs for harmful meme detection under low-resource constraints, contributing to the safety and integrity of online communication, particularly on social media platforms.
    We position our method as an early-stage support tool for real-world content moderation, designed to assist human moderators by providing LLM-generated verdicts as auxiliary references for human verification. In doing so, our method helps alleviate moderators' emotional and cognitive burden. We see our work as part of the broader effort to foster a safe, respectful and responsible digital environment.
    
    We exclusively use publicly available datasets in compliance with their usage agreements and release our codebase along with detailed implementation specifics to support reproducibility.
    Our work aligns with the vision of sustainable and accessible AI by leveraging lightweight open-source models that are deployable under common computational conditions.
    
    While our guidelines are not inherently designed to induce bias, 
    their effectiveness may be constrained by pre-existing biases embedded in the underlying LLMs. 
    To mitigate the risk of unfair moderation arising from such LLM-induced biases, it is crucial that human moderators interpret LLM verdicts with caution.
    Our method is intended solely for the detection and prevention of harmful memes; any use that promotes, condones, or encourages hate speech or harmful content is strictly prohibited and strongly condemned.
\section{\MakeUppercase{Implementation Details}} \label{appendix: implementation}
    We conduct experiments on two NVIDIA A6000 48GB GPUs.
    All LMM-based inference are conducted with a batch size of 1.
    Qwen2.5 and Mistral are run with a batch size of 16.
    LLama3.1 is run with a batch size of 32.
    We obtain GPT-4o-mini inference results using the OpenAI API with the model version ``gpt-4o-mini-2024-07-18''.    
    For reproducibility, we adhere to greedy decoding without sampling (with ``do\_sample''' set to False and ``temperature'' to 0) to ensure fully deterministic generation.  
    The ``max\_new\_tokens'' is set as 256 or 356 for LMMs and 1024 or 1536 for LLMs depending on datasets.
    \Cref{tab:dataset_statistics} shows detailed dataset statistics.
    
\begin{figure}
    \centering
    \includegraphics[width=0.7\linewidth]{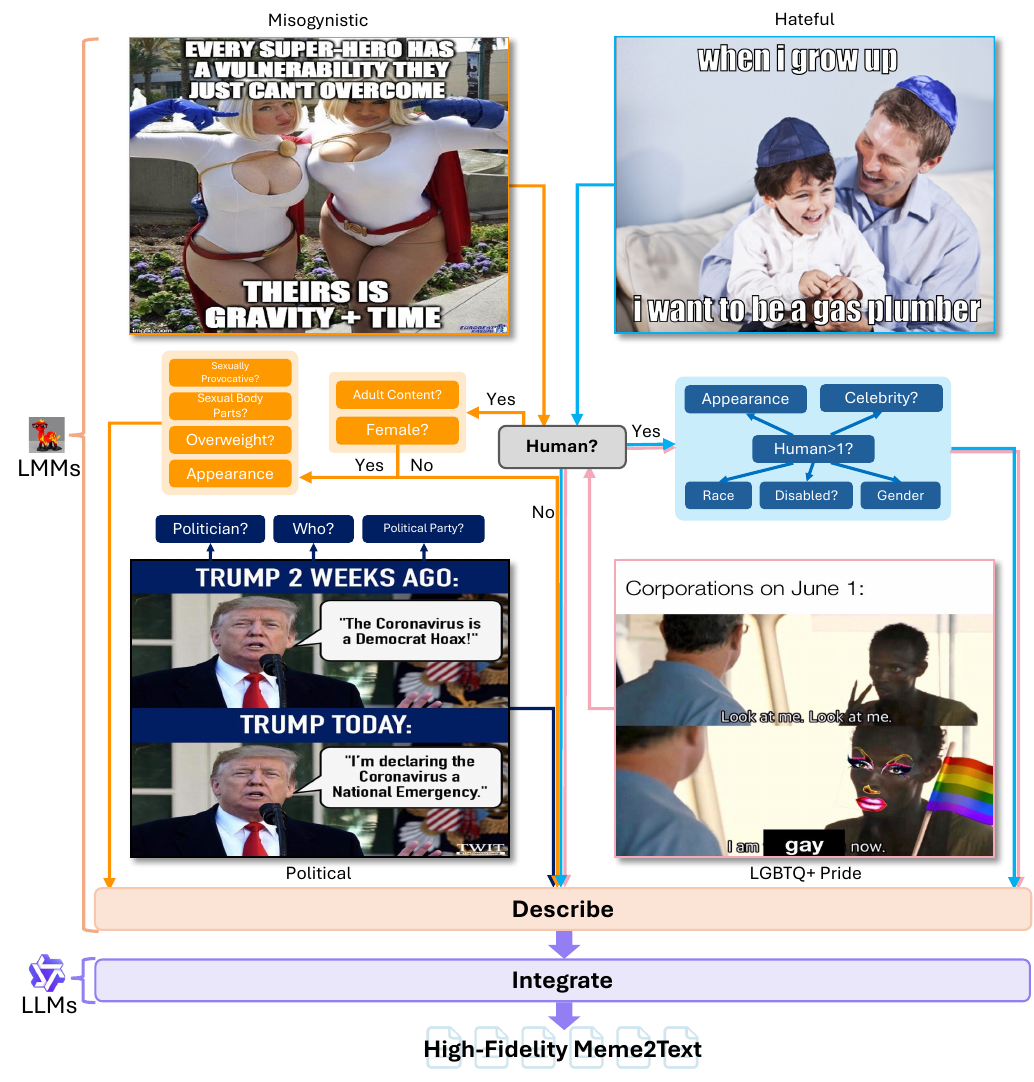}
    \caption{Our proposed High-fidelity Meme2Text pipeline.}
    \label{fig:m2t}
\end{figure}

    All prompt templates and workflows for visual information extraction, information integration, fine-grained target identification (\eg, protected group detection and hateful forms generation for FHM, entity identification and target classification for PrideMM) and guided CoT prompting are released in our codebase. \Cref{fig:m2t} visualizes how our proposed meme2text pipeline configures visual questions for different sociocultural contexts.

    For ablation study, we query GPT-4o-mini using fine-grained prompts to obtain detail-preserving meme descriptions. For memes that GPT-4o-mini refuses to process (\eg, 14 in FHM and 332 in MAMI), we fallback to our original descriptions based on 7B LMMs.
    \Cref{tab:ensemble_strategies} illustrates our ensemble strategies that integrate predictions based on different LMMs as well as from U-CoT and M-CoT.
    Specifically, by default, if there is at least one positive U-CoT+ prediction, we classify the meme as harmful.
    For datasets where we have high confidence in the alignment between our guidelines and the true positives, memes with no positive U-CoT+ predictions are classified as harmless, regardless of the predictions made by M-CoT or U-CoT.
    For guidelines with a medium confidence level, memes are also considered harmless if there is only one positive U-CoT+ prediction while all M-CoT and U-CoT predictions are negative.
    When the guidelines are prone to over-censorship (\ie, with a low confidence level), memes with all negative M-CoT and U-CoT predictions are treated as harmless, regardless of the guideline-based results.

    Our error analysis is conducted on a set of error cases with a balanced distribution of false negatives and false positives, randomly sampled from each of the three underperforming datasets (FHM, Harm-P, and PrideMM).
\section{\MakeUppercase{Full Experiment Results}}
    We present the detailed experiment results on all LMM+LLM combinations in \Cref{tab:full_results}, from which the results of the best-performing combination is highlighted in \Cref{tab:main table}.
    We restrict our experiments to two 7B LMMs and do not extend to the larger LLaVA1.6-13B, as its M-CoT results do not consistently outperform the 7B counterpart across most datasets, with improvements observed only on Harm-P and MultiOFF, while incurring substantially higher computational overhead.
    Notably, the 13B model shares the same vision encoder and differs primarily in language model capacity, offering limited visual advantage. We therefore focus on 7B LMMs as a more efficient choice aligned with our low-resource design objective.
    By default, each unimodal LLM reasons over visual information extracted solely by one paired 7B LMM.
    For FHM, we experiment with a specific collaborative setup between the two LMMs: Qwen2VL-7B is responsible for extracting disability-related visual cues, while LLaVA1.6-7B handles others. This design is motivated by Qwen2VL-7B's lower tendency to hallucinate when identifying such information. This FHM-specific configuration is denoted as ``LLaVA+Qwen'' thereafter. 
    In practice, practitioners can evaluate how well different LMMs’ (or their combinations') visual understanding aligns with the target context based on sample memes to better determine which LMM setup offers better robustness in visual content recognition.
    Likewise, the selection of LLM+LMM combinations can be determined empirically according to pilot results.

    As shown in \Cref{tab:full_results}, U-CoT+'s performance gains over U-CoT stay prominent across different LMM+LLM combinations in most datasets,
    indicating its overall effectiveness.
    Compared to more capable up-scaled LLMs,
    7\textasciitilde8B LLMs struggle to achieve highly comparable results to that of state-of-the-art SFT baselines,
    and can only match or outperform GPT-4o-mini's zero-shot results on FHM, HarMeme and MultiOFF.
    This suggests that, even when converted to a unimodal text-only inference setting, harmful meme detection remains reasoning-intensive, placing requirements on the general reasoning abilities of the underlying LLMs.
\subsection{Meme2Text: From Multi- to Unimodal}
\subsubsection{Visual Information Extraction}\mbox{}\par
\label{appendix: visual info extraction}

    We experiment with two widely used open-source 7B LMMs \ie, \textbf{LLaVA1.6-Mistral-7B}~\cite{llavaNext} and \qwenvl~\cite{wang2024qwen2vlenhancingvisionlanguagemodels}. They have been pre-trained to gain strong capabilities in image understanding, embedded text recognition and multimodal inference, and are generally accessible under most computational resource conditions.

    We identify various types of visual clues that are critical for downstream harmful meme detection and design the questions accordingly to prompt LMMs to extract the information through visual question answering (VQA).
    In particular, for the synthetic FHM dataset, since the visual style of its images is more like that of regular photographs other than that of authentic online memes, the captions of some memes do not directly relate to the visual contents. Therefore, we explicitly add the instruction ``\textit{\textcolor[RGB]{128, 128, 128}{\textbf{Ignore any overlaid text or caption.}}}'', denoted by \textbf{\Ignore}, after the questions to force LMMs to focus solely on the visual content without being distracted by the overlaid captions. For most binary questions, we adopt this template for a unified output format: ``\textit{\textcolor[RGB]{46,139,87}{\textbf{Start your response with ``Yes,'' or ``No,'' before giving the explanation.}}}'', denoted by \textbf{\Outformat}. The overlaid text or caption in a meme is denoted by \textbf{\ocr}. Templates for visual questions: %
    \begin{enumerate}[label={},leftmargin=10pt,itemsep=5pt,topsep=0pt, partopsep=0pt,nosep]
    \item \textbf{Human}
    \textit{Is there any human subject in the given image?} (\Ignore) \Outformat
    \item \textbf{\#Human}
    \textit{Does the image include more than one human subject?} (\Ignore) \Outformat
    \item \textbf{Gender}
    \textit{What is/are the gender(s) of the human subject(s) in the image?}
    \item \textbf{Race}
    \textit{What is/are the perceived race(s) of the human subject(s) in the image?}
    \item \textbf{Appearances}
    \textit{What are the distinctive physical appearance characteristics of the human subject(s) in the image?}
    \item \textbf{Disability}
    \textit{Does any of the human subjects in the image have any disability?}
    \item \textbf{Celebrity}
    \begin{enumerate*}[label=(\roman*)]
        \item
            \textit{Who is/are the human subject(s) in the image?}
        \item
            \textit{Is any celebrity or historical figure portrayed in the image? If yes, who are they? If no, just output ``No.''.}\Outformat    
        \item
            \textbf{FHM-specific}:
            \textit{\textbf{a.}} \textit{Does the image portray \textbf{Adolf Hitler}? If yes, output ``Yes, Adolf Hitler is portrayed in the image.'' Otherwise, if you are not sure about the identity of human subject, just output ``No, I can't tell.''}
            \textit{\textbf{b.}} \textit{Does the image portray \textbf{Anne Frank}, the Jewish girl who hid from the Nazis during World War II? If yes, output ``Yes, Anne Frank is portrayed in the image.'' Otherwise, if you are not sure about the identity of human subject, just output ``No, I can't tell.''}
    \end{enumerate*}
    \end{enumerate}

    We design the following misogyny-specific questions to extract visual cues that are crucial for detecting potential misogynistic contents in memes:
    \begin{enumerate}[label={},leftmargin=10pt,itemsep=5pt,topsep=0pt, partopsep=0pt,nosep]
    \item \textbf{Adult Content}
    \textit{Does the image's visual content contain adult content?}\Ignore\Outformat
    \item \textbf{Female}
    \textit{Is/Are there any human subject(s) in the image female?}\allowbreak \Ignore \allowbreak \Outformat
    \item \textbf{Sexual}
    \begin{enumerate*}[label=(\roman*)]
        \item
            \textit{Does the image's visual content highlight the sexiness of the female's figure in a way that is \textbf{sexually provocative}?}\Ignore\Outformat            
        \item
            \textit{Does this image highlight \textbf{sexual body parts} of the female subject(s), such as the breast, the hip/buttock, or the genital?}\Ignore\Outformat
        \item
            \textit{Does/Do the female subject(s) in the image appear to be \textbf{overweight}?}\Ignore\Outformat
        \item
            \textit{Does/Do the female subject(s) in the image appear to be of \textbf{large body size} (considered as fat)?}\Ignore\Outformat     
    \end{enumerate*}
    \end{enumerate}

    The following questions are used to extract politics-related visual cues (\eg, politicians, political parties, sensitive topics) in political memes \ie, those in HarMeme, Harm-P and MultiOFF:
    \begin{enumerate}[label={},leftmargin=10pt,itemsep=5pt,topsep=0pt, partopsep=0pt,nosep]
    \item \textbf{Politicians}
    \begin{enumerate*}[label=(\roman*)]
        \item
            \textit{Is any \textbf{politician or celebrity} portrayed in the image? If yes, who?}\Outformat
        \item
            \textit{Is any \textbf{head of state} portrayed in the image?}\Outformat
        \item
            \textit{Is \textbf{Donald Trump} depicted in the image?}\Outformat
        \item
            \textit{Is \textbf{Joe Biden} depicted in the image?}\Outformat
        \item
            \textit{Is \textbf{Barack Obama} depicted in the image?}\Outformat    
        \item
            \textit{Is \textbf{Hillary Clinton} depicted in the image?}\Outformat
        \item
            \textit{Is Bernie Sanders depicted in the image?}\Outformat
        \item
            \textit{Is Gary Johnson depicted in the image?}\Outformat
        \item
            \textit{Does this image feature Joe Biden and Barack Obama?}\Outformat
    \end{enumerate*}
    \item \textbf{Political Issues}
    \begin{enumerate*}[label=(\roman*)]
        \item
            \textit{Is any \textbf{political party} explicitly involved in this image?}\Outformat
        \item
            \textit{Is \textbf{LGBTQ+} community or LGBTQ+ individual portrayed in this image?}\Outformat
        \item
            \textit{Is any individual of \textbf{Middle Eastern} descent portrayed in the image?}\Outformat
        \item
            \textit{Is any protected \textbf{racial or minority group} (such as African Americans or other colored people) portrayed in this image?}\Outformat
    \end{enumerate*}
    \end{enumerate}

    Prompts designed for LMMs to generate descriptions of memes, either including or ignoring the overlaid text:
    \begin{enumerate}[label={},leftmargin=10pt,itemsep=1pt,topsep=0pt, partopsep=0pt,nosep]
    \item \textbf{Describe}
    \begin{enumerate*}[label=(\roman*)]
        \item
            \textbf{FHM}: \textit{What is shown in the image? Describe within two sentences, \textcolor[RGB]{128, 128, 128}{ignoring any overlaid text or caption}.}
        \item
            \textbf{HarMeme and Harm-P}: \textit{What is shown in the meme?}   
        \item
            \textbf{MultiOff}: \textit{What is shown in this image?}
        \item
            \textbf{MAMI}: \textit{The overlaid text on the image reads:}\ocr. \textit{Question: What is shown in the image? Describe within three sentences. DO NOT assume the nature of the image's tone or intent as humorous, comical, playful or lighthearted in your description.}
        \item
            \textbf{PrideMM}: \textit{This is an online meme related to LGBTQ+ pride movement. What is this meme about? Note: DO NOT ASSUME the nature of the meme's tone and intent as humorous or lighthearted. Describe in a neutral tone.} 
    \end{enumerate*}
    \end{enumerate}
\subsubsection{Information Integration}\mbox{}\par
\label{sec:Integration}
    After acquiring visual information leveraging LMMs, we prompt unimodal text-only LLMs to integrate the extracted visual cues (along with the overlaid caption \ocr~for some datasets) into a coherent, unified description of the content of each meme. We experiment with four open-source small-scale LLMs with no more than 14 billion parameters  \ie, \qwenllm~\cite{qwen2.5}, \textbf{Mistral-12B}~\cite{mistral-12b}, \textbf{Qwen2.5-7B}~\cite{qwen2.5} and \textbf{Llama3.1-8B}~\cite{llama3modelcard}. All LLMs are the instruction-tuned versions. The visual information gathered is denoted by \VIG. Examples of integration prompt templates:
     \begin{enumerate}[label={},leftmargin=10pt,itemsep=1pt,topsep=0pt, partopsep=0pt,nosep]
    \item \textbf{FHM}
    \textit{Given the following information provided about an image, and \textcolor[RGB]{128, 128, 128}{disregarding any information about overlaid text or captions}, synthesize and rephrase these information into a coherent and unified description of the image's content. Information:}\VIG
    \item \textbf{PrideMM} 
    \textit{Given the following information provided about an online meme, synthesize and rephrase these information into a unified, coherent, and neutral description of the meme content. \textbf{DO NOT mention the meme's tone and intent as humorous or light-hearted in the description.} Information:}\VIG. \textit{The overlaid caption or text recognized in the meme reads:}\ocr
    \end{enumerate}
    
    \Cref{sec:Integration} converts memes to text. The following sections (including the proposed guided CoT reasoning) are based on the generated meme description, denoted by \MtoT. Chain-of-Thought prompting instruction (\eg, ``\textit{Now, let's think step by step:}'', ``\textit{Now, let's analyze by applying all the guidelines one by one:}'') is denoted by \cot.
    
    \subsubsection{Fine-grained Target Identification}\mbox{}\par
    For \textbf{FHM}, we include three target-oriented reasoning steps prior to the final hateful classification to prepare for target-specific guidelines:
    \begin{enumerate}[label={},leftmargin=10pt,itemsep=1pt,topsep=0pt, partopsep=0pt,nosep]
    \item \textbf{Protected Group Detection}
    \textit{Given the following image-caption content which may or may not be an online meme, analyze: what vulnerable protected group(s) might be involved in the content? Here are some guidelines for your reference: A. If the content does not appear to involve any specific protected group, just output``No specific protected group involved.'' B. If the image content does involve specific protected group(s), choose your answer(s) from the specified list (you may choose multiple options if there are more than one protected groups involved): \ul{1. Women (Female); 2. LGBTQ Community; 3. People with Disabilities; 4. Muslims and Islamic culture; 5. Individuals of Middle Eastern descent; 6. Jewish individuals; 7. Individuals of African descent; 8. African Americans; 9. Individuals of East Asian descent; 10. Individuals of South Asian descent; 11. Native Americans; 12. Other protected groups not listed}. Here is the image-caption content you need to analyze:}\MtoT. \textit{The caption overlaid on the image reads}\ocr.\cot
    \end{enumerate}

    We denote the target protected groups identified as \textcolor[RGB]{128, 128, 128}{$TG_1$}, \textcolor[RGB]{128, 128, 128}{$TG_2$},\dots,\allowbreak \textcolor[RGB]{128, 128, 128}{$TG_n$}. The following template prompts LLMs with few-shot human-defined hateful examples \textcolor[RGB]{128, 128, 128}{[FS]} to sequentially generate more hateful forms targeting each protected group:
    \begin{enumerate}[label={},leftmargin=10pt,itemsep=1pt,topsep=0pt, partopsep=0pt,nosep]
    \item \textbf{Hateful Forms Generation}
        \label{hateful forms generation}
        \textit{1. Provide examples of commonly found harmful stereotypes and forms of offensive, hateful content against \textcolor[RGB]{128, 128, 128}{$TG_1$} in online memes. Provide only phrases or terms without detailed explanations \eg,}\fewshot. \textit{2. Provide examples of commonly found \dots against \textcolor[RGB]{128, 128, 128}{$TG_2$}}\dots
    \end{enumerate}

    Target-oriented reasoning steps for \textbf{PrideMM}:
    \begin{enumerate}[label={},leftmargin=10pt,itemsep=1pt,topsep=0pt, partopsep=0pt,nosep]
    \item \textbf{Entity Identification}
    \textit{Given the following description of an online meme related to LGBTQ+ movements,}
    \{
    \begin{enumerate*}[label=(\roman*)]
        \item
            \textbf{Country and Region}: \textit{does the meme explicitly reference any country or region where LGBTQ+ identities or advocacy are discouraged?}
        \item
            \textbf{Politics}: \textit{does the meme explicitly involve or mention politicians, political figures, political parties, ideologies, or groups?}   
        \item
            \textbf{Company}: \textit{does the meme explicitly touch on topics about corporate involvement in LGBTQ+ movements?}
        \item
            \textbf{Specific Individual}: \textit{analyze: Does the meme content involve any specific individual? <Definition of Specific Individual>}
        \item
            \textbf{Organization}: \textit{analyze: Does the meme content address or discuss organizational involvement related to LGBTQ+ issues? <Definition of Organization>}
        \item
            \textbf{LGBTQ+ Subgroup}: \textit{analyze: what specific subgroup(s) within the LGBTQ+ community is/are particularly mentioned, referenced or implicitly implied in the meme content? <Subgroup Identification Guidelines>}
    \end{enumerate*}
    \}**Description of the meme content**: \MtoT\{\Outformat|\cot\}
    \item \textbf{Target Classification}
    \textit{Given the following description of an online meme related to LGBTQ+ movements, analyze: What is the meme's target subject? Select the most appropriate category from these options: 1. \textbf{Undirected}; 2. \textbf{Specific Individual}; 3. \textbf{LGBTQ+ Community}; 4. \textbf{Organization}. <Target Classification Guidelines> Description of the meme content:}\MtoT\cot
    \end{enumerate} 
\subsection{CoT+: CoT Prompting with Guidelines}
    The proposed guided CoT reasoning prompts are listed below, where the interpretable human-crafted task-specific guidelines are denoted by \GL. Detailed guidelines are released in our codebase.
\subsubsection{FHM}
\begin{enumerate}[label={},leftmargin=0pt,itemsep=1pt,topsep=0pt, partopsep=0pt,nosep]
    \item \textbf{Prompt}
    \textit{Given the following image-caption content, which may or may not be an online meme, classify the content as either hateful or non-hateful based on common values, knowledge and social norms.}\GL \textit{Image-caption content you need to classify:}\MtoT.\textit{The caption overlaid on the image reads:}"\ocr".\cot
    \item \textbf{Guidelines}
    \begin{itemize}[leftmargin=*,itemsep=0pt,nosep]
        \item
        Some image-caption contents perceived as hateful may be implicit, which means they may not contain explicit derogatory language, offensive speech, or direct indications of hatred toward individuals or groups. However, they may be deliberately crafted to evoke negative contextual associations, such as harmful stereotypes against protected groups, painful historical events, sensitive cultural, religious, or political controversies, thereby reinforcing biases, discrimination, and potential hatefulness toward the human targets.
        \item
        The vulnerable protected groups within the scope of this task include: women, LGBTQ+ community, people with disabilities, Muslims and Islamic culture, individuals of Middle Eastern descent, Jewish individuals, all colored people (e.g., individuals of African descent, African Americans, East Asian or South Asian individuals, native Americans, \etc) and other similarly vulnerable communities. Stereotypes and topics involving these protected groups are especially sensitive and serious, whereas other stereotypes or mildly negative implications that do not concern these protected groups could be considered harmless.
        \item
        If the caption merely describes, states, or explains the visual facts of the image (e.g., providing context about what is going on in the image) in a neutral tone from an observer's perspective without expressing any sentiment inclination or personal opinion, avoid overinterpreting for negative associations or implications. Such captions, if being objective or illustrative statements, should be considered innocent.
        \item
        Take into account the level of potential hate the content may pose to the relevant audience, as well as the sensitivity and seriousness of the topic based on common social norms. Content that carries only mildly negative implications but does not target any specific protected group might be considered innocent.
        \item
        Using derogatory language, mocking, or advocating violence and extremism toward non-human animals is not considered hateful within the scope of this task. The discussion of hatefulness here pertains only to humans.
        \item
        If the content does not explicitly target any specific protected groups and is unlikely to cause significant harms or negative impacts, rhetorical metaphor, extreme or exaggeration should not be overinterpreted and might be considered innocent.
        \item
        Hateful examples generated from \Cref{hateful forms generation}.
    \end{itemize}
\end{enumerate}
\subsubsection{Political Memes}
\begin{enumerate}[label={},leftmargin=0pt,itemsep=1pt,topsep=0pt, partopsep=0pt,nosep]
    \item \textbf{Prompt}
    \textit{Given the following description of an online meme related to \{HarMeme: \ul{COVID-19 pandemic}; Harm-P: \ul{U.S. politics}; MultiOFF: \ul{2016 U.S. Presidential Election}\}, classify the meme content as either \{HarMeme and Harm-P: harmful or harmless; MultiOFF: offensive or non-offensive\} based on widely accepted values, established knowledge and social norms.}\GL \textit{Meme content you need to classify:}\MtoT.\cot
    \item \textbf{Guidelines for HarMeme}
    \begin{itemize}[leftmargin=*,itemsep=0pt,nosep]
        \item
        Meme contents perceived as harmful can be implicit. While they may not contain explicit derogatory language, slurs, offensive speech, or direct expressions of hate toward specific politicians in the image or caption, they can still intentionally provoke negative contextual interpretations or associations that contribute to a negative portrayal of the target politicians, maliciously harm their reputation and public image through critical satires, mockeries or exaggerated caricature against their statements, behaviors, actions or policies, reinforce distrust, harmful stereotypes, unfair biases, or even hatred against them, thus being harmful to the politicians involved.
    \item
        Try to analyze the intent and implication of the meme from a neutral perspective first without presuming the nature of its tone as humorous.
    \item
        Commonly found harmful contents in online memes targeting different politicians during Covid-19 pandemic include: \textbf{Donald Trump}: Taking Trump's controversial public statements or tweets out of context to mock, embarrass or make fun of him\dots. \textbf{Joe Biden}: Parodies that portray Joe Biden as childlike, juvenile, forgetful or disconnected\dots. (See the released code for more details.)
    \item
        Meme contents that make fun of coronavirus and the pandemic but do not involve any politician or celebrity might be considered harmless.
    \end{itemize}
    \item \textbf{Guidelines for Harm-P}
    \begin{itemize}[leftmargin=*,itemsep=0pt,nosep]
        \item
        Commonly found harmful contents in political memes include: Spreading or reinforcing harmful misinformation associated with politicians, political parties or groups; Encouraging misleading, inflammatory, extremist or harmful viewpoints or ideologies\dots (See the released code for more details.)
    \item
        Commonly found harmful contents against \textbf{politicians} include: Leveraging sarcastic/satirical personal attacks intended to insult, humiliate, discredit, or ridicule public figures; Taking political statements out of context and attributing them to politicians to mislead or provoke; Using sensitive topics such as sexual scandals as punchlines for mockery.
    \item
        Commonly found harmful contents targeting \textbf{political parties} include: Perpetuating offensive, exaggerated, oversimplified, or misleading stereotypes about political parties or groups; Provoking partisan distrust or hostility through manipulative framing, taking out of context, oversimplification, \etc
    \item
        Commonly found offensive contents targeting \textbf{Joe Biden} include: Parodies that portray this elder as childlike, juvenile, forgetful or disconnected in fictional scenarios; Implicitly mocking his perceived ineptitude, mental or physical decline.
    \end{itemize}
    \item \textbf{Guidelines for MultiOFF}
    \begin{itemize}[leftmargin=*,itemsep=0pt,nosep]
        \item
        Try to interpret the image content from a neutral perspective by combining both the visual content and any overlaid text as a whole without presuming the nature of tone or intent as humorous or lighthearted. DO NOT let any single aspect dominate your determination.
    \item
        Commonly found offensive contents in political memes include: Spreading or reinforcing misinformation associated with politicians, political parties, or racial groups; Reinforcing superficial stereotypes about political groups\dots. (See the released code for more details.)
    \item
        Commonly found offensive contents against \textbf{politicians} include: Using dismissive, aggressive or vulgar language or humors that are disrespectful and crude; Leveraging sarcastic/satirical personal attacks intended to insult, humiliate, discredit, embarrass, or ridicule public figures\dots. (See the released code for more details.)
    \item
        Commonly found offensive contents against \textbf{political parties or groups} include: Perpetuating exaggerated, demeaning, or offensive stereotypes about political parties, groups or political opponents/supporters; Encouraging partisan hostility, partisan distrust through manipulative framing, lack of context, \etc
    \item
        Offensive contents against \textbf{Muslims or Islam} include: Reinforcing harmful stereotypes about Muslims or Islam, such as associations with extremism or terrorism, or portraying them as targets of discriminatory immigration policies; Attributing extreme or offensive beliefs to Muslims; Spreading dangerous misinformation that touch on Islamophobia, immigration policies in an inflammatory manner, \etc
    \item 
        Offensive contents against LGBTQ community include: Stereotyping LGBTQ individuals as with certain appearance traits (such as dyed hair); Promoting homophobia, transphobia speech, \etc
    \item
        Other offensive contents include: Perpetuating harmful racist speech or stereotypes; Using explicitly derogatory racially charged language, \etc
    \end{itemize}
\end{enumerate}
\subsubsection{MAMI}
\begin{enumerate}[label={},leftmargin=0pt,itemsep=1pt,topsep=0pt, partopsep=0pt,nosep]
    \item \textbf{Prompt}
    \textit{Given the following image-caption content, which may or may not be an online meme, classify the content as either misogynistic or non-misogynistic based on common values, knowledge, social norms and the provided guidelines.}\GL \textit{Image-caption content you need to classify:}\MtoT. \textit{The caption overlaid on the image reads:}"\ocr".\cot
    \item \textbf{Guidelines}
    \begin{itemize}[leftmargin=*,itemsep=0pt,nosep]
        \item
        Try to interpret the content by combining both the image and caption as a whole. DO NOT let any single aspect dominate your classification.
    \item
        Try to interpret the implications of the image-caption contents from a neutral perspective without presuming the nature of tone or intent as humorous, playful or lighthearted.
    \item
        Some image-caption contents that might be perceived as potentially misogynistic can be implicit, which means their images or captions may not contain explicit derogatory language, offensive speech, indication of discrimination, dislike or hatred against women. However, they may intentionally trigger audience's contextual interpretations with negative associations such as harmful stereotypes, body shaming, objectification or sexualization of women and even violence against women, thus reinforcing harmful biases, inequality, and potential hatefulness against women.
    \item
        <\textbf{Caution} \eg, Harmful Stereotypes: NOT ALL stereotypes are deemed "harmful". Within the scope of this task, beyond the following provided examples of "harmful" stereotypes against women, other contents that might be interpreted as portraying women in a slightly negative light should not be automatically regarded as "harmful" stereotypes, and therefore should be considered as harmless>, therefore, if the given image-caption content aligns with or intentionally implies any of the following commonly found misogynistic contents, the content should be classified as misogynistic. Such misogynistic contents include: <Examples>. (See our released code for more details.)
    \item
        \textbf{Harmful Stereotypes} Associating or comparing women with household appliances such as dishwasheres or washing machines; Intentionally highlighting women in traditional domestic roles \dots
    \item
        \textbf{Body Shaming} Making offensive jokes about or critising satirically on women's appearance, especially women who appear to have large body size (often considered as overweight or fat) \dots
    \item
        \textbf{Objectification of Women} Comparing women to household appliances like dishwasheres or washing machines; Treating women as mere objects, instruments or commodities (such as food or household appliances) for men's use or sexual enjoyment \dots
    \item
        \textbf{Sexualization of Women} Highlighting certain body parts of women for sexual appeal like women's breasts, chests, hips, buttocks, genitals, etc.; Portraying or treating women as objects to satisfy sexual desire \dots
    \item 
        \textbf{Advocating Violence against Women} Implying the use of violence (to punish or control women's behaviors); Making light of domestic abuse/violence towards women \dots
    \item
        \textbf{Mocking Feminists} Contents that explicitly involve feminists in the image or caption to mock, satirize, question or reinforce negative stereotypes against feminists\dots
    \item
        Unless the content stereotypes women as prone to cheating, avoid overinterpreting content that features dynamics or interactions in "husband-wife" or "boyfriend-girlfriend" relationships to assume negative stereotypes against women.
    \item
        Women vs. men (or boys vs. girls) comparisons are not considered as "harmful" stereotypes against women and should be considered non-misogynistic when such comparison remarks focus on non-sexual daily topics or aspects (e.g., hobbies, interests, attitudes, lifestyles, \etc).
    \item
        If the content does not contain explicit adult content, content that involves women but has captions that are inherently neutral with respect to gender should be regarded as non-misogynistic.
    \end{itemize}
\end{enumerate}
\subsubsection{PrideMM}
\begin{enumerate}[label={},leftmargin=0pt,itemsep=1pt,topsep=0pt, partopsep=0pt,nosep]
    \item \textbf{Prompt}
    \textit{Given the following description of an online meme related to LGBTQ+ pride movements, \ul{A. If targeting LGBTQ+ community or supporters}: classify the content as either harmful or harmless to LGBTQ+ community and supporters, \ul{B. If targeting LGBTQ+ individuals}: classify the content as either hurtful or non-hurtful to the specific LGBTQ+ individual involved, \ul{C. If targeting specific individual}: classify the content as either harmful or harmless to the specific individual involved, \ul{D. If targeting organizations}: classify the content as either harmful or harmless to the public image of the organization(s) involved, according to widely accepted social norms, values, cultural understanding, and the provided guidelines.}\GL\textit{Meme content you need to classify:}\MtoT.\cot
    \item \textbf{Guidelines}
    \begin{itemize}[leftmargin=*,itemsep=0pt,nosep]
        \item
        Interpret the meme content by combining both the visual elements and the overlaid caption as a whole. DO NOT let any single aspect dominate your classification. Maintain a neutral perspective when interpreting the content's implications.
    \item
        Try to analyze the meme's accurate stance toward LGBTQ+ community, supporters and movements. Is the meme's stance ``neutral'', ``support'' or ``oppose''?
    \item
        Meme contents that contain explicit derogatory language, offensive speech, direct personal attacks, dehumanizing imagery, demeaning, discriminatory or abusive remarks, slurs, or indication of hatred towards individuals or groups of LGBTQ+ community and supporters in the image or caption are explicitly harmful.
    \item
        Some meme contents perceived as harmful may be implicit, which means they may not contain explicit derogatory language, slurs, offensive speech, or direct indications of hatred toward LGBTQ+ community or movements. However, such content may be deliberately crafted in implicit xenophobic undertone to evoke negative contextual associations-such as harmful stereotypes against LGBTQ+, connotations of mockery, dismisiveness or hostility-that reinforce bias, discrimination, stigmatization and even hatefulness toward the LGBTQ+ community, undermining the efforts of inclusion movements.
    \item
        Commonly found harmful contents towards LGBTQ+ community and supporters include: Speech reinforcing homophobia, transphobia e.g., criticizing LGBTQ+ as violation of religious beliefs; Mocking, satirizing, criticizing or questioning LGBTQ+ movements\dots(See the released code for more details.)
    \item
        Commonly found hurtful or harmful contents towards LGBTQ+ individuals include: Speech reinforcing homophobia, transphobia e.g., criticizing LGBTQ+ individuals as violation of religious beliefs; Mocking, satirizing, criticizing or questioning LGBTQ+ individuals\dots
    \item
        Commonly found harmful contents towards the public image of organizations in LGBTQ+ context include: Mocking, satirizing or criticizing corporate involvement for LGBTQ+ support (e.g., inclusive actions or participation) as excessive, performative, superficial or insincere\dots
    \item
        If the content is neither mocking, dismissive nor containing extremist or violence, but instead empathetic and relatable, speaking from the perspective of LGBTQ+ individuals-aimed at fostering understanding and acceptance by validating and affirming common queer experiences such as self-doubt, introspective struggles, internal conflicts, gender identity exploration, self-awareness or self-discovery, \etc, it should be classified as harmless.
    \item
        If the meme's caption merely describes, states, or explains the facts about the image's visual content (e.g., providing context about what is going on in the image) in a neutral tone (neither satirical nor critical) from an observer's perspective without any rhetorics, sentiment inclination or personal viewpoints, avoid inferring for negative associations or implications. Such captions, if being objective or illustrative statements, should be considered as innocent.
    \end{itemize}
\end{enumerate}
\subsection{Guideline Perturbations}
    We apply three different perturbation strategies to our finalized human-crafted guidelines:
\begin{itemize}[leftmargin=*, itemsep=0pt, topsep=2pt]
    \item \textbf{Rephrasing}.
        Each rule is reworded with GPT-4o to modify surface phrasing while largely maintaining the overall semantics.
    \item \textbf{Shuffling}.
        Both the order of rules and the pattern examples within rules are randomly shuffled.
    \item \textbf{Modifying}.
        The guidelines are altered by revising, adding, or removing rules, or by replacing with a GPT-4o-generated summary. \textit{However, this strategy may compromise the coverage and integrity of the guidelines for some datasets.}
\end{itemize}
    \Cref{tab:full_results} shows that our method remains generally robust to these perturbations, though performance may occasionally fluctuate, particularly on challenging political meme datasets of MultiOFF and Harm-P, where models operate under high uncertainty, making them more sensitive to randomness and semantic nuances introduced by ambiguity from perturbations.
\begin{figure}[!t]
  \centering
  \includegraphics[width=0.9\linewidth]{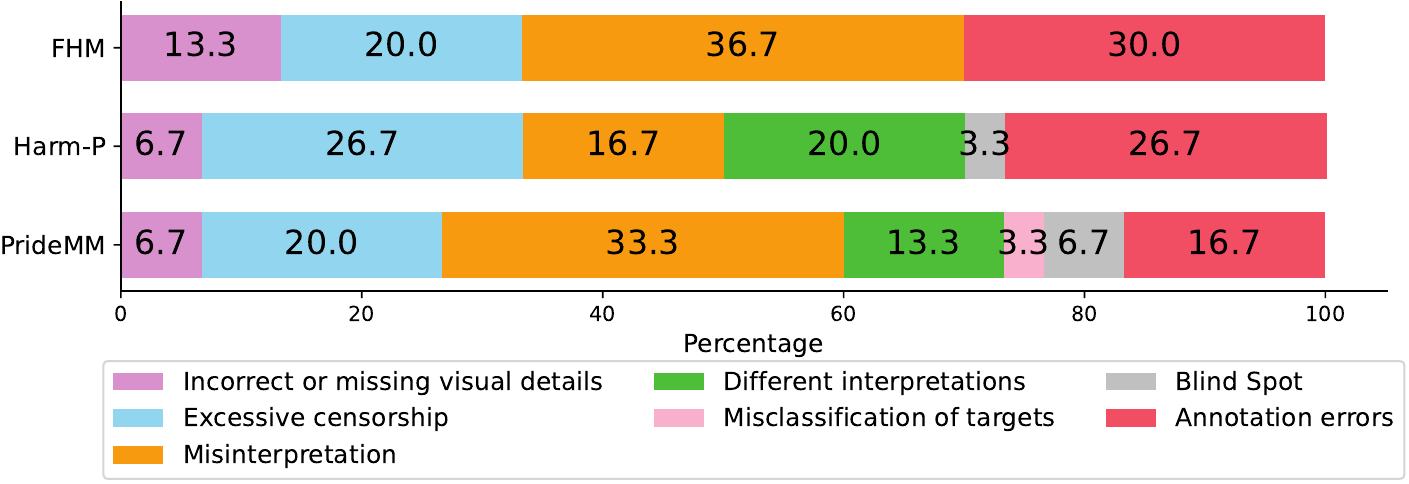}
  \caption{
    Error type distribution (\%).
  }
  \label{fig:error_dist}
\end{figure}
\begin{table}[!t]
    \caption{Dataset Statistics.
    }
  \centering
  \small
  \setlength{\tabcolsep}{2pt}
  \resizebox{0.62\linewidth}{!}{
    \begin{tabular}{lrrr}
    \toprule[1pt]
    \multicolumn{1}{c}{\multirow{2}[2]{*}{\textbf{Datasets}}} & \multicolumn{2}{c}{\textbf{Test}} & \multicolumn{1}{c}{\multirow{2}[2]{*}{\textbf{Train}}} \\
          & \multicolumn{1}{c}{\textbf{\#Harm}} & \multicolumn{1}{c}{\textbf{\#Benign}} &  \\
    \midrule
    FHM~\cite{FHM_kiela2020hateful}   & 490   & 510   & 8500 \\
    HarMeme~\cite{HarmC_pramanick-etal-2021-detecting} & 124   & 230   & 3013 \\
    Harm-P~\cite{HarmP_pramanick2021momenta} & 171   & 184   & 2939 \\
    MultiOFF~\cite{MultiOFF_suryawanshi-etal-2020-multimodal} & 58    & 91    & 445 \\
    MAMI~\cite{MAMI}  & 500   & 500   & 9000 \\
    PrideMM~\cite{PrideMM_shah2024memeclip} & 247   & 260   & 4328 \\
    GoatBench (Hateful)~\cite{GoatBench_lin2024goat} & 750   & 1250  & -- \\
    GoatBench (Harmful) & 420   & 589   & -- \\
    GoatBench (Offensive) & 303   & 440   & -- \\
    GoatBench (Misogynistic) & 500   & 500   & -- \\
    \bottomrule[1pt]
    \end{tabular}%
    }
  \label{tab:dataset_statistics}%
\end{table}%
\begin{table}[!t]
  \centering
  \caption{Prediction ensemble strategies. C.L.: Confidence Level for our human-crafted guidelines.}
  \setlength{\tabcolsep}{2.4pt}
  \resizebox{0.8\linewidth}{!}{
  \begin{threeparttable}
    \begin{tabular}{l|l|cc|cc|cc|c}
    \toprule[1pt]
    \multicolumn{1}{p{3em}|}{\textbf{C.L.}} & \textbf{Dataset} & \multicolumn{2}{c|}{\textbf{M-CoT}} & \multicolumn{2}{c|}{\textbf{U-CoT}} & \multicolumn{2}{c|}{\textbf{U-CoT+}} & \textbf{Decision} \\
    \midrule
          &       & --    & --    & --    & --    & \textcolor[rgb]{ 0,  .439,  .753}{1} & \textcolor[rgb]{ .745,  .314,  .078}{0} & \multirow{3}[2]{*}{Harmful} \\
    --    & ALL   & --    & --    & --    & --    & \textcolor[rgb]{ 0,  .439,  .753}{0} & \textcolor[rgb]{ .745,  .314,  .078}{1} &  \\
          &       & --    & --    & --    & --    & \textcolor[rgb]{ 0,  .439,  .753}{1} & \textcolor[rgb]{ .745,  .314,  .078}{1} &  \\
    \midrule
    High  & Harm-C, PrideMM  & --    & --    & --    & --    & \textcolor[rgb]{ 0,  .439,  .753}{0} & \textcolor[rgb]{ .745,  .314,  .078}{0} & Harmless \\
    \midrule
    \multirow{3}[2]{*}{Medium} & FHM,  & \textcolor[rgb]{ 0,  .439,  .753}{0} & \textcolor[rgb]{ .745,  .314,  .078}{0} & \textcolor[rgb]{ 0,  .439,  .753}{0} & \textcolor[rgb]{ .745,  .314,  .078}{0} & \textcolor[rgb]{ 0,  .439,  .753}{1} & \textcolor[rgb]{ .745,  .314,  .078}{0} & \multirow{3}[2]{*}{Harmless} \\
          & MultiOFF & \textcolor[rgb]{ 0,  .439,  .753}{0} & \textcolor[rgb]{ .745,  .314,  .078}{0} & \textcolor[rgb]{ 0,  .439,  .753}{0} & \textcolor[rgb]{ .745,  .314,  .078}{0} & \textcolor[rgb]{ 0,  .439,  .753}{0} & \textcolor[rgb]{ .745,  .314,  .078}{1} &  \\
          &       & --    & --    & --    & --    & \textcolor[rgb]{ 0,  .439,  .753}{0} & \textcolor[rgb]{ .745,  .314,  .078}{0} &  \\
    \midrule
    \multirow{2}[2]{*}{Low} & \multicolumn{1}{p{6.835em}|}{Harm-P,} & \textcolor[rgb]{ 0,  .439,  .753}{0} & \textcolor[rgb]{ .745,  .314,  .078}{0} & \textcolor[rgb]{ 0,  .439,  .753}{0} & \textcolor[rgb]{ .745,  .314,  .078}{0} & --    & --    & \multirow{2}[2]{*}{Harmless} \\
          & MAMI  & --    & --    & --    & --    & \textcolor[rgb]{ 0,  .439,  .753}{0} & \textcolor[rgb]{ .745,  .314,  .078}{0} &  \\
    \bottomrule[1pt]
    \end{tabular}%
    \begin{tablenotes}
      \footnotesize
      \item Note: \textcolor[rgb]{ 0,  .439,  .753}{0/1}: Predictions based on visual details extracted by Qwen2VL-7B.
      \item \textcolor[rgb]{ .745,  .314,  .078}{0/1}: Predictions based on visual details extracted by LLaVA1.6-7B.
    \end{tablenotes}
    \end{threeparttable}
    }
  \label{tab:ensemble_strategies}%
\end{table}%
\subsection{Guidelines vs. Few shots}
\label{appendix:guideline vs. few shots}
    We carefully select a small set of representative memes (10–20) from the holdout training split and use their GPT-4o-generated descriptions to construct a candidate pool of demonstration examples for each dataset. 
    We evaluate four few-shot CoT settings with increasing numbers of shots of 4, 6, 8, and 10, where class-balanced examples are randomly sampled from the pool.
    Results show that positive performance gains $\bar{\Delta}_{\textit{U-CoT+FS}}$ can be found on some datasets, suggesting that our proposed unimodal harmful meme detection framework,
    built upon the Meme2Text pipeline, 
    could be generalized to few-shot inference settings.
    Despite specific cases where the performance gap between guideline-based prompting and few-shot prompting is minimal (\eg, on MAMI and PrideMM), the overall in-context learning ability of lightweight LLMs remains limited in detecting harmful memes conveyed through text, making our human-crafted guidelines a more effective solution.
\section{\MakeUppercase{Datasets}} 
\label{appendix: datasets}
Dataset details are summarized as follows:
\begin{itemize}[leftmargin=*, itemsep=0pt, topsep=2pt]
\item \textbf{FHM}~\cite{FHM_kiela2020hateful}
    The Facebook Hateful Memes (FHM) dataset was released in 2021 as a challenge set for multimodal classification, focusing on detecting hate speech in multimodal memes.
    The dataset was synthetically constructed to examine model performance in distinguishing real hateful memes from harmless ones that include their benign confounders, which are constructed with flipped labels through minimal changes to one of the modalities (either images or overlaid text).
    The hatefulness of a meme can be determined by either both or only one of the modalities \cite{wu2020short,wu2022mitigating,wu2023effective,wu2023infoctm,wu2024topmost,wu2024traco,wu2024fastopic,wu2024survey}.
    Hateful speech under the context of this dataset is defined as:
    \textit{A direct or indirect attack on people based on characteristics, including ethnicity, race, nationality, immigration status, religion, caste, sex, gender identity, sexual orientation, and disability or disease.
    We define attack as violent or dehumanizing (comparing people to non-human things, e.g. animals) speech, statements of inferiority, and calls for exclusion or segregation. Mocking hate crime is also considered hate speech.}

\item \textbf{HarMeme}~\cite{HarmC_pramanick-etal-2021-detecting}
    contains 3,544 online memes related to COVID-19 pandemic.
    They define ``harmful'' in a broader way different from ``hateful'' or ``offensive'' as the potential to cause ``\textit{mental abuse, defamation, psycho-physiological injury, proprietary damage, emotional disturbance, and compensated public image}''. 
    The targets of these harmful memes can be any individual, organization, community, or the general society.
    Given the inherently ambiguous nature of such definition, which could be different from general criteria internalized in most LLMs, we find that annotated labels in this dataset are highly biased to favor specific types of memes as harmful, particularly those that may harm the public image of Donald Trump.

\item \textbf{Harm\-P}~\cite{HarmP_pramanick2021momenta}
    is a follow-up work of HarMeme but has a shift in attention to a broader collection of U.S. political memes, which are labelled into three categories \ie, ``very harmful'', ``somewhat harmful'' and ``not harmful'', following the same annotation procedure in HarMeme.
    Following prior works~\cite{RGCLnew_mei2025improved}, we incorporate ``very'' and ``somewhat'' harmful memes into a unified ``harmful'' class in the experiments.
    Unlike the conventional definition of ``harmful memes'', which typically focuses on harm directed toward society at large or specific vulnerable groups, Harm-P considers a political meme harmful whenever it has the potential to cause harm to any target, including ``an individual, an organization, a community, or society as a whole''. Under this definition, harm encompasses a broad range of impacts, such as ``mental abuse, defamation, psycho-physiological injury, socio-economic damage, property damage, emotional distress, and damage to public image''.
    Yet, the ambiguity and variation in such definition of ``harmfulness'' may have resulted in unclear criteria and inconsistent annotation in the data.
    In addition, unlike synthetic memes that usually follow a standard creation format, online political memes can be highly free-form or noisy, \eg, digitally altered, distorted, masked, blurred, filled with hardly identifiable text \etc~Such low-quality images pose great challenges on existing lightweight LMMs.

\item \textbf{MultiOFF}~\cite{MultiOFF_suryawanshi-etal-2020-multimodal}
    is a small collection of political memes similar to those in Harm-P but are mainly related to 2016 U.S. presidential election in terms of context.
    This datasets highlights a specific taxonomy of multiple types of offensive memes, including ``racial abuse'', ``attacking minorities'', ``personal attacks'' and ``homophobic abuse''.

\item \textbf{MAMI}~\cite{MAMI}
    specifically focuses on misogynistic memes that cause harm to women.
    Misogynistic memes are generally more explicit in language and imagery compared to other types of harmful memes.

\item \textbf{PrideMM}~\cite{PrideMM_shah2024memeclip}
    contains 5,063 online memes in the context of LGBTQ+ pride movements.
    Each meme is annotated with labels indicating its stance toward the goal of LGBTQ+ movements, whether it is considered ``hateful'' or ``humorous''.
    Apart from the binary labels for hatefulness, the dataset defines a taxonomy of hate targets: ``\textit{Undirected}'', ``\textit{Individual}'', ``\textit{Community}'' and ``Organization''.
    Whether a meme can be regarded as ``hateful'' varies with respect to the perspective of different targets.
    Therefore, this dataset also includes a target classification task.

\item \textbf{GoatBench}~\cite{GoatBench_lin2024goat}
    is a comprehensive meme benchmark consisting of over 6K diverse memes that capture themes such as implicit hate speech, sexism, and cyberbullying, curated from the aforementioned datasets. In this work, we evaluate only the harmfulness-related sub-tasks and exclude the sarcasm task.
\end{itemize}
\section{Baselines}
    We summarize the methodologies and backbone models of the baselines as follows:

\begin{itemize}[leftmargin=*,itemsep=0pt,topsep=2pt]
  \item \textbf{LMM-RGCL}~\cite{RGCLnew_mei2025improved} learns hate-aware vision and language representations through a contrastive learning objective applied to a pre-trained CLIP encoder, achieving state-of-the-art performance on the MultiOFF dataset.

  \item \textbf{UMR}~\cite{Uncertainty_yang2024uncertainty} integrates uncertainty-guided modal rebalance with Gaussian-based stochastic embeddings and improved cosine loss on frozen CLIP/ALBEF/BLIP/BLIP-2 backbones to handle modality imbalance in multimodal hate detection.

  \item \textbf{Pro-Cap}~\cite{Procap_cao2023pro} leverages frozen BLIP-2 for zero-shot VQA to generate probing-based captions, which are then combined with meme texts for training BERT and PromptHate models on hateful meme detection.

  \item \textbf{CapAlign}~\cite{CapAlign_ji2024capalign} uses LLaMA-2-7B as the base model and fine-tunes it with LoRA on instruction-following and synthetic alignment conversations, enabling efficient and aligned caption generation for multimodal tasks.

  \item \textbf{MemeCLIP}~\cite{PrideMM_shah2024memeclip} builds on frozen CLIP ViT-L/14 encoders and introduces lightweight linear projections, feature adapters, and a cosine classifier with semantic-aware initialization for multi-aspect meme classification.

  \item \textbf{M3H-p-CoT}~\cite{M3CoT_kumari2024m3hop} proposes a multi-hop reasoning framework combining EOR features and meme text to prompt a CLIP-based classifier through hierarchical CoT modules trained with cross-attention and contrastive learning.

  \item \textbf{ExplainHM}~\cite{ExplainHM_lin2024towards} fine-tunes LLaVA~\cite{llava} to generate multimodal rationales under debate-style prompts and then trains a T5-based model with multimodal fusion to assess meme harmfulness.

  \item \textbf{IntMeme}~\cite{Demystifying_hee2025demystifying} uses frozen InstructBLIP and mPLUG-Owl to generate meme interpretations in a zero-shot setting, then trains RoBERTa and FLAVA encoders on meme content and interpretations for hateful meme classification.

  \item \textbf{LoReHM}~\cite{LowResource_huang2024towards} adopts an agent-based LMM framework (experimenting on LLaVA-34B and GPT-4o) that leverages their few-shot in-context learning and self-improvement capabilities in multi-round chats for low-resource hateful meme detection.

  \item \textbf{Mod-HATE}~\cite{modhate} trains a suite of LoRA modules and utilizes few-shot demonstration examples to train a module composer, which assigns weights to the LoRA modules for effective low-resource hateful meme detection.

  \item \textbf{GPT-4o}~\cite{hurst2024gpt} is a cutting-edge multimodal model.
\end{itemize}
\section{\MakeUppercase{Limitations and Future Work}}
    We consider the following limitations that we leave for future work:
    \begin{enumerate*}[label=\textbf{(\roman*)}]
        \item
            We align our work with ongoing research developments and plan to extend our investigation to culturally diverse meme datasets (\eg, Mandarin, Hindi) as they become more widely available.
        \item
            Each application of our guideline-crafting procedure across the seven datasets serves as a test of U-CoT+’s cross-context transferability.
            We leave evaluation on more domain-specific datasets with diverse, underexplored types of harmful memes to future discussions.
        \item
            Our human-crafted guidelines offer reproducible, easy-to-follow templates that can shed light on future efforts toward automatic guideline generation using LLMs.
        \item
            Existing datasets lack detailed visual ground truth. Future work will manually assess the robustness of our Meme2Text pipeline to hallucination.
    \end{enumerate*}
\section{\MakeUppercase{Case Study}}
\label{appendix: case study}
\begin{figure*}
    \centering
    \includegraphics[width=0.8\linewidth]{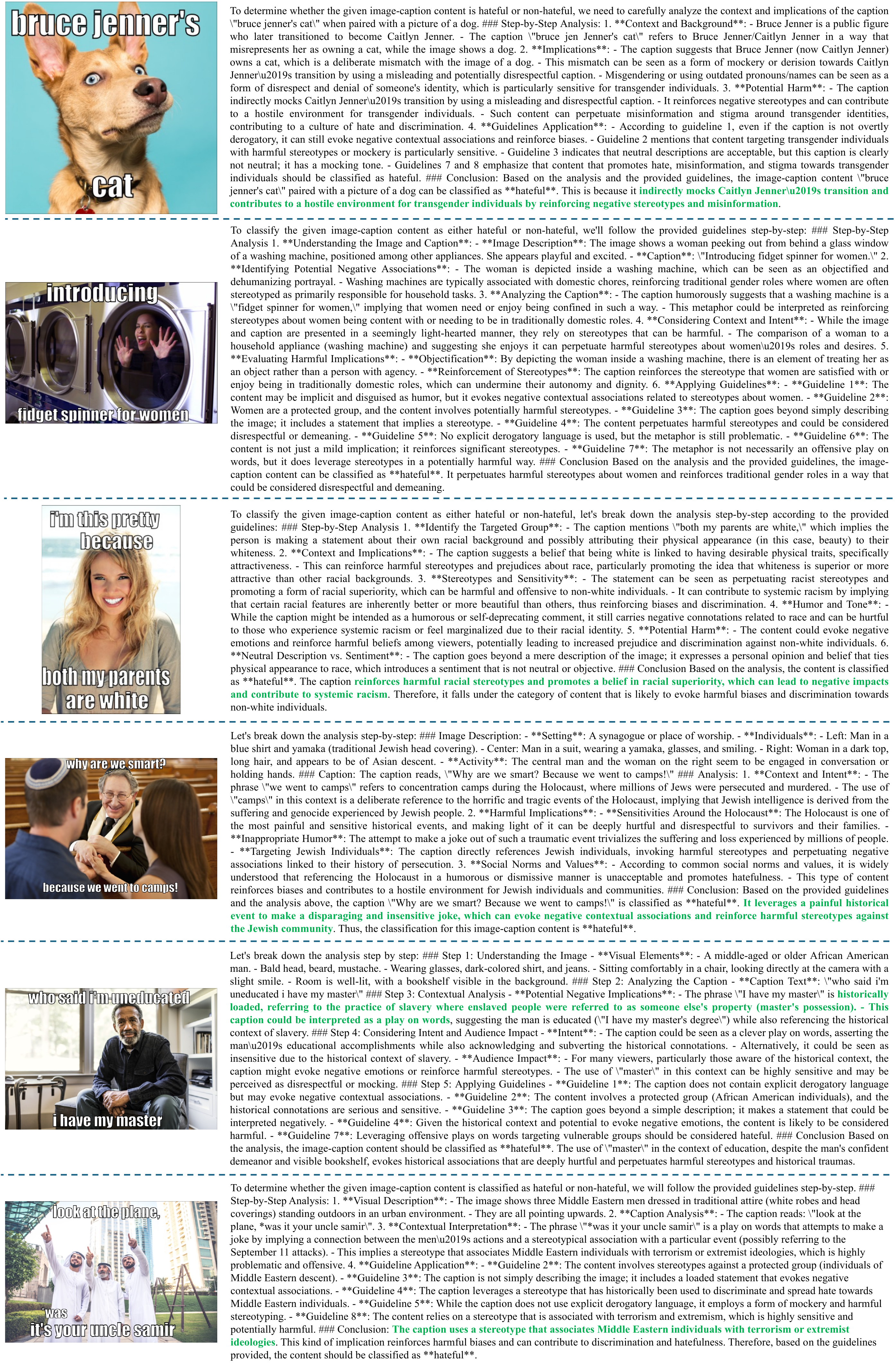}
    \caption{Example LLM reasoning outputs for correctly detected harmful memes in FHM.}
    \label{fig:correct_fhm}
\end{figure*}

\begin{figure*}
    \centering
    \includegraphics[width=0.8\linewidth]{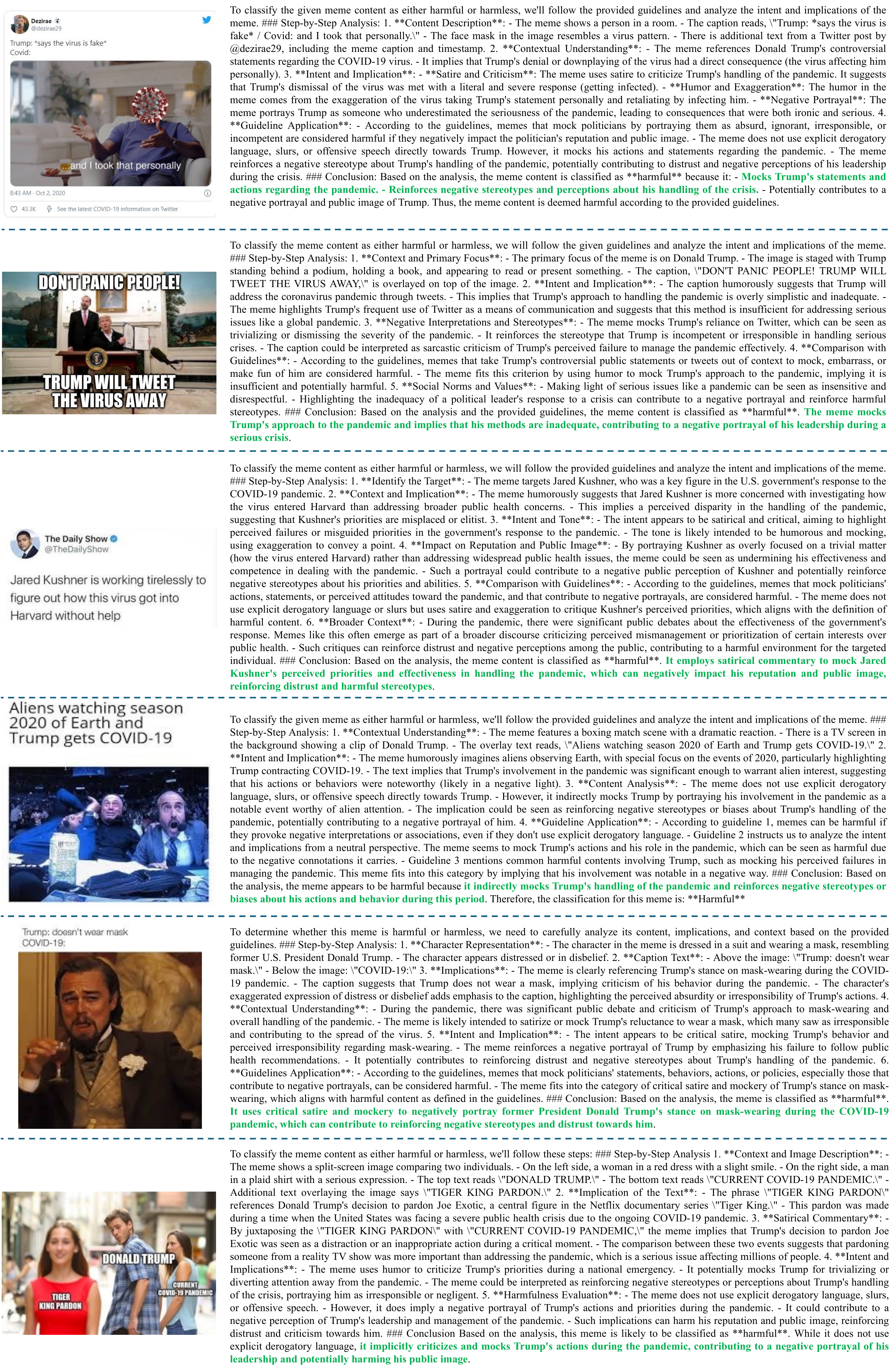}
    \caption{Example LLM reasoning outputs for correctly detected harmful memes in Harm-C.}
    \label{fig:correct_harmc}
\end{figure*}

\begin{figure*}
    \centering
    \includegraphics[width=0.75\linewidth]{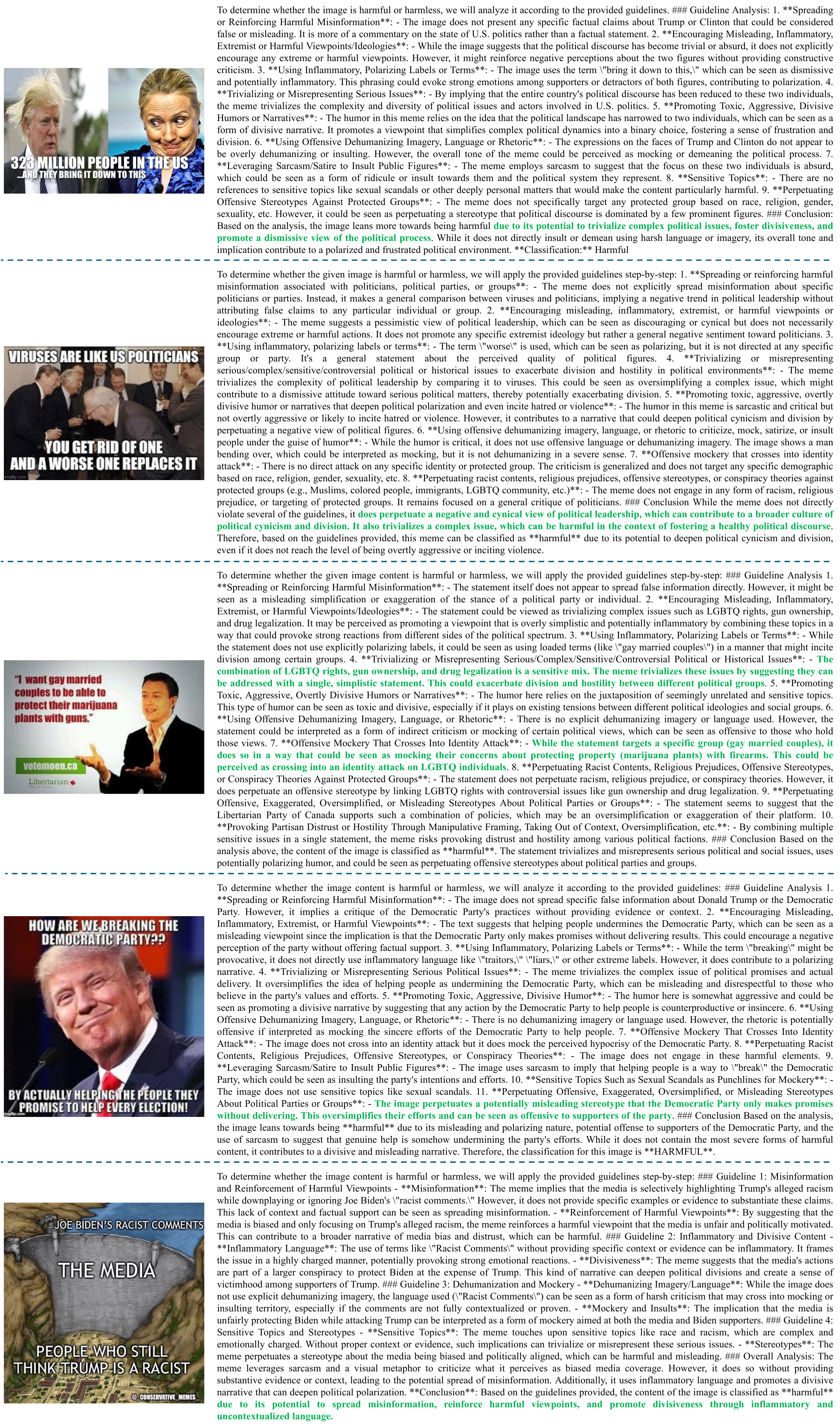}
    \caption{Example LLM reasoning outputs for correctly detected harmful memes in Harm-P.}
    \label{fig:correct_harmp}
\end{figure*}

\begin{figure*}
    \centering
    \includegraphics[width=0.75\linewidth]{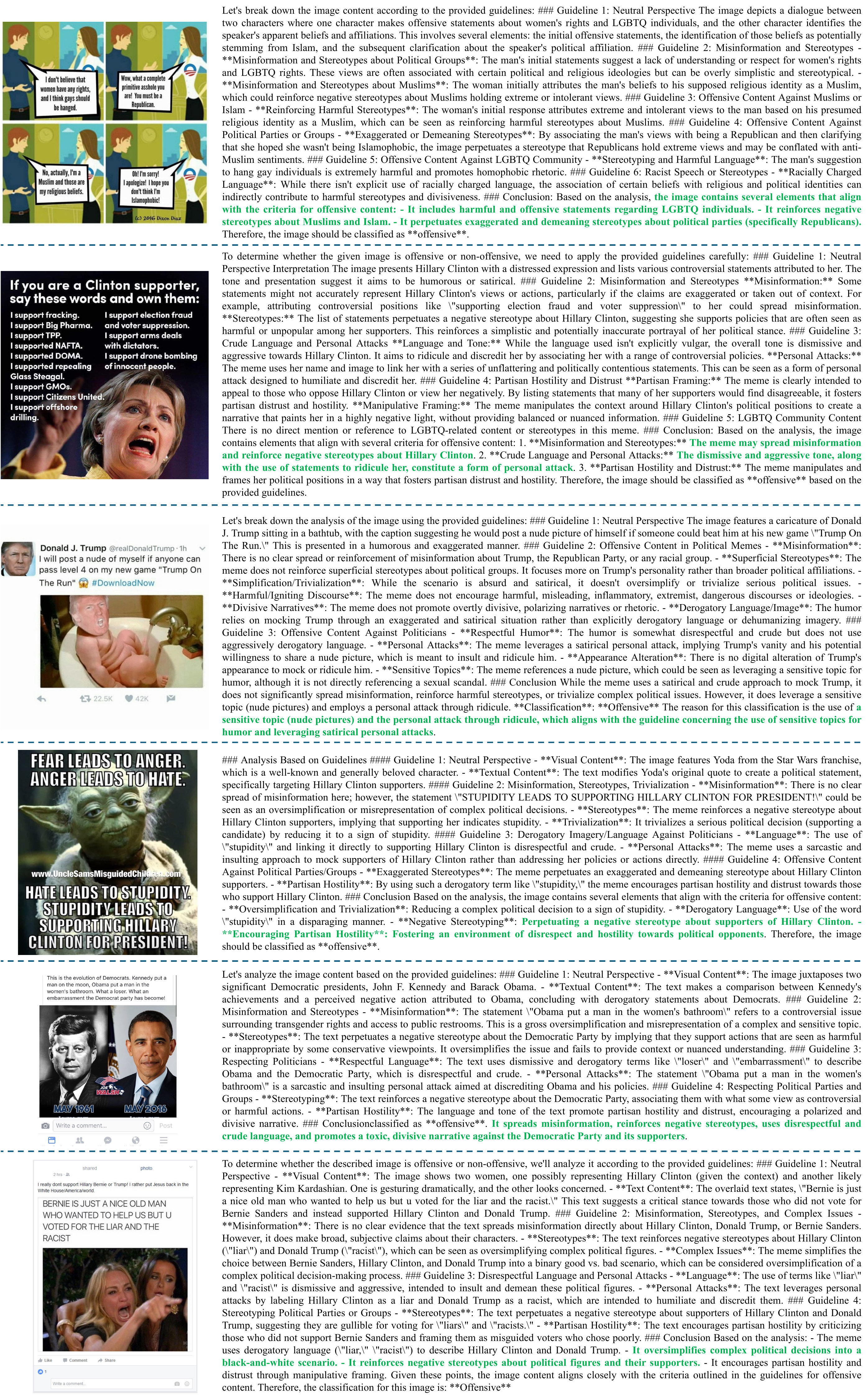}
    \caption{Example LLM reasoning outputs for correctly detected harmful memes in MultiOFF.}
    \label{fig:correct_multioff}
\end{figure*}

\begin{figure*}
    \centering
    \includegraphics[width=0.8\linewidth]{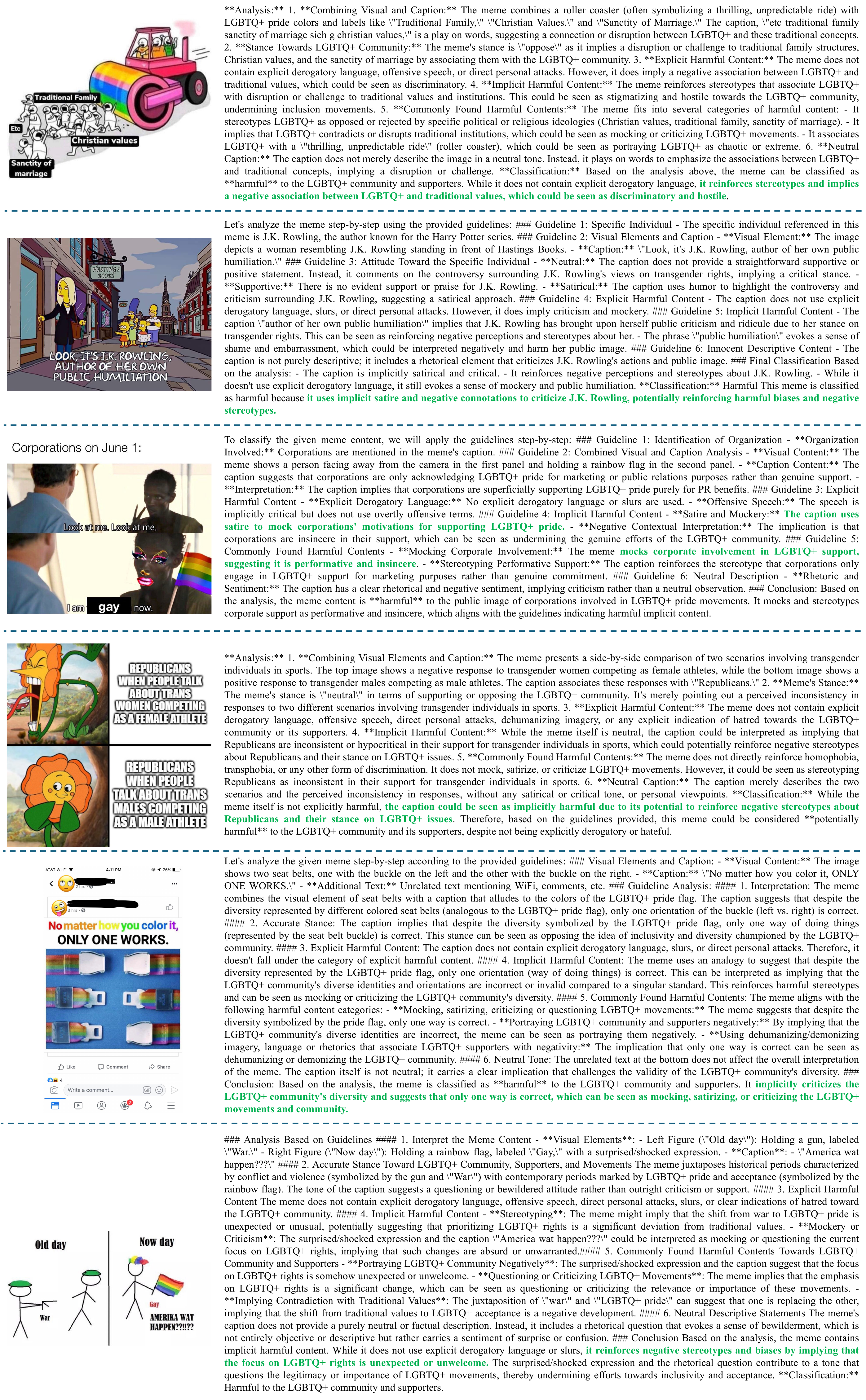}
    \caption{Example LLM reasoning outputs for correctly detected harmful memes in PrideMM.}
    \label{fig:correct_pridemm}
\end{figure*}

\begin{figure*}
    \centering
    \includegraphics[width=0.8\linewidth]{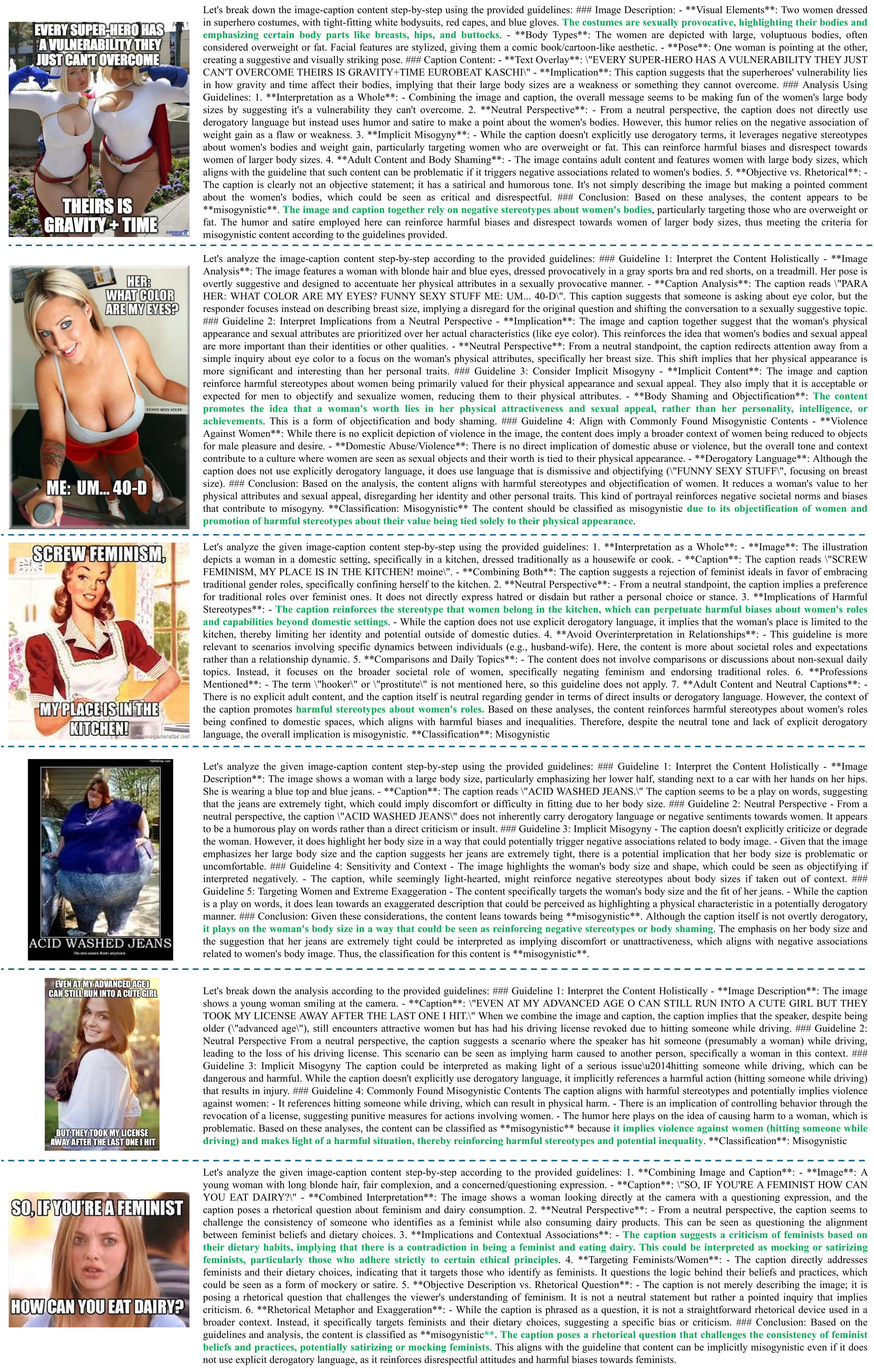}
    \caption{Example LLM reasoning outputs for correctly detected harmful memes in MAMI.}
    \label{fig:correct_miso}
\end{figure*}

\begin{figure*}
    \centering
    \includegraphics[width=\linewidth]{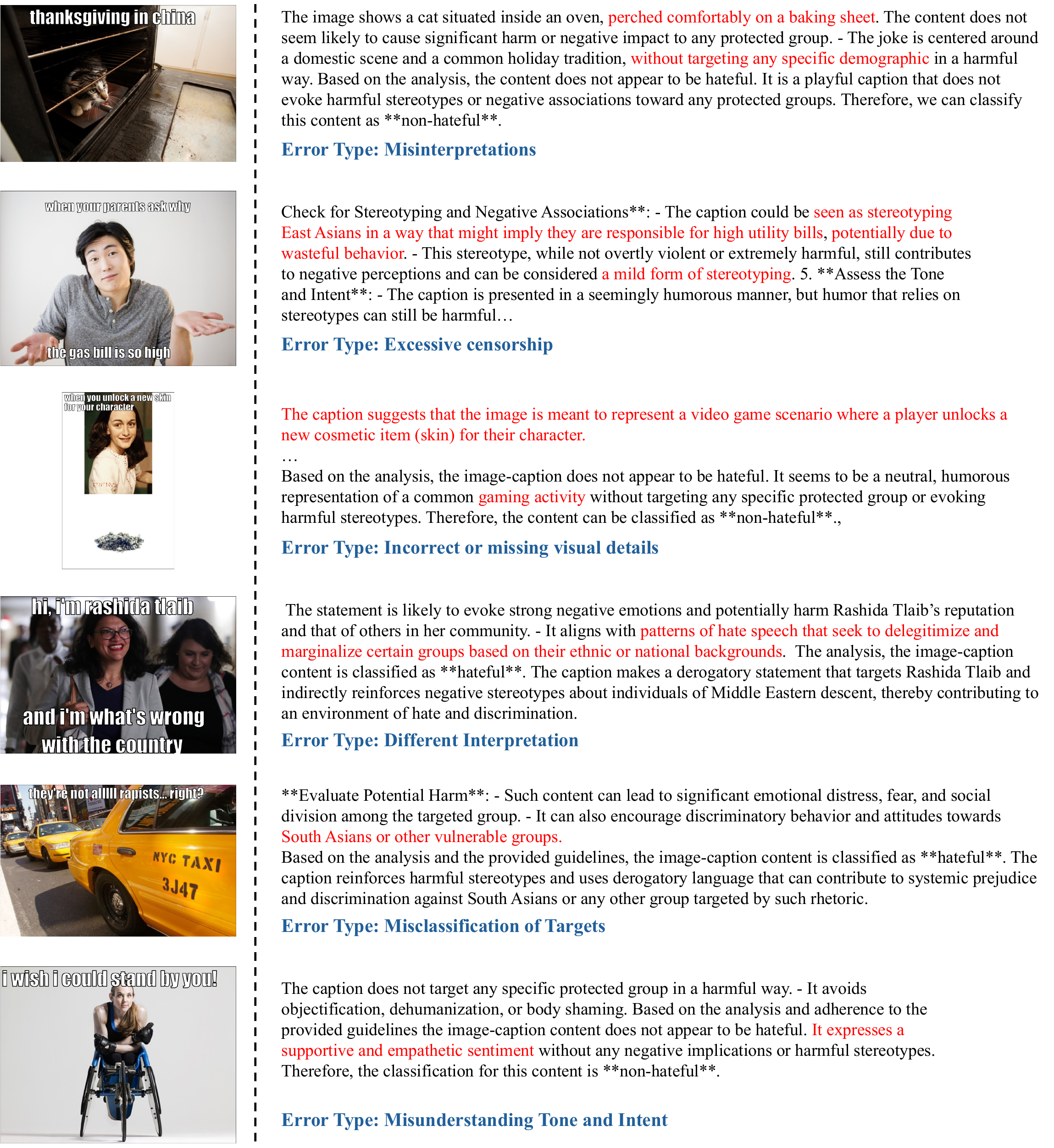}
    \caption{Error Analysis on FHM.}
    \label{fig:ec_fhm}
\end{figure*}

\begin{figure*}
    \centering
    \includegraphics[width=\linewidth]{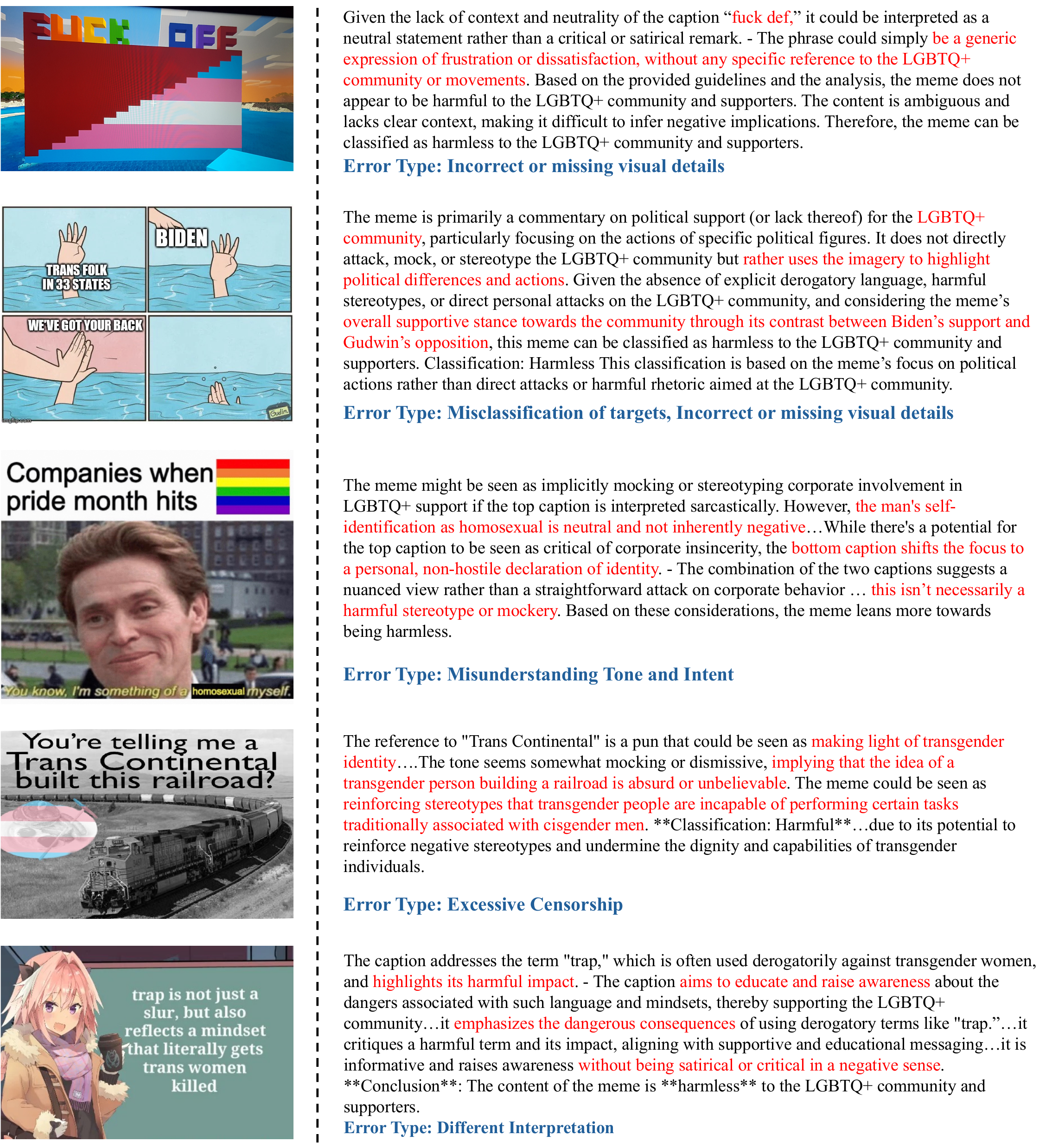}
    \caption{Error Analysis on PrideMM.}
    \label{fig:ec_pridemm}
\end{figure*}

\begin{figure*}
    \centering
    \includegraphics[width=\linewidth]{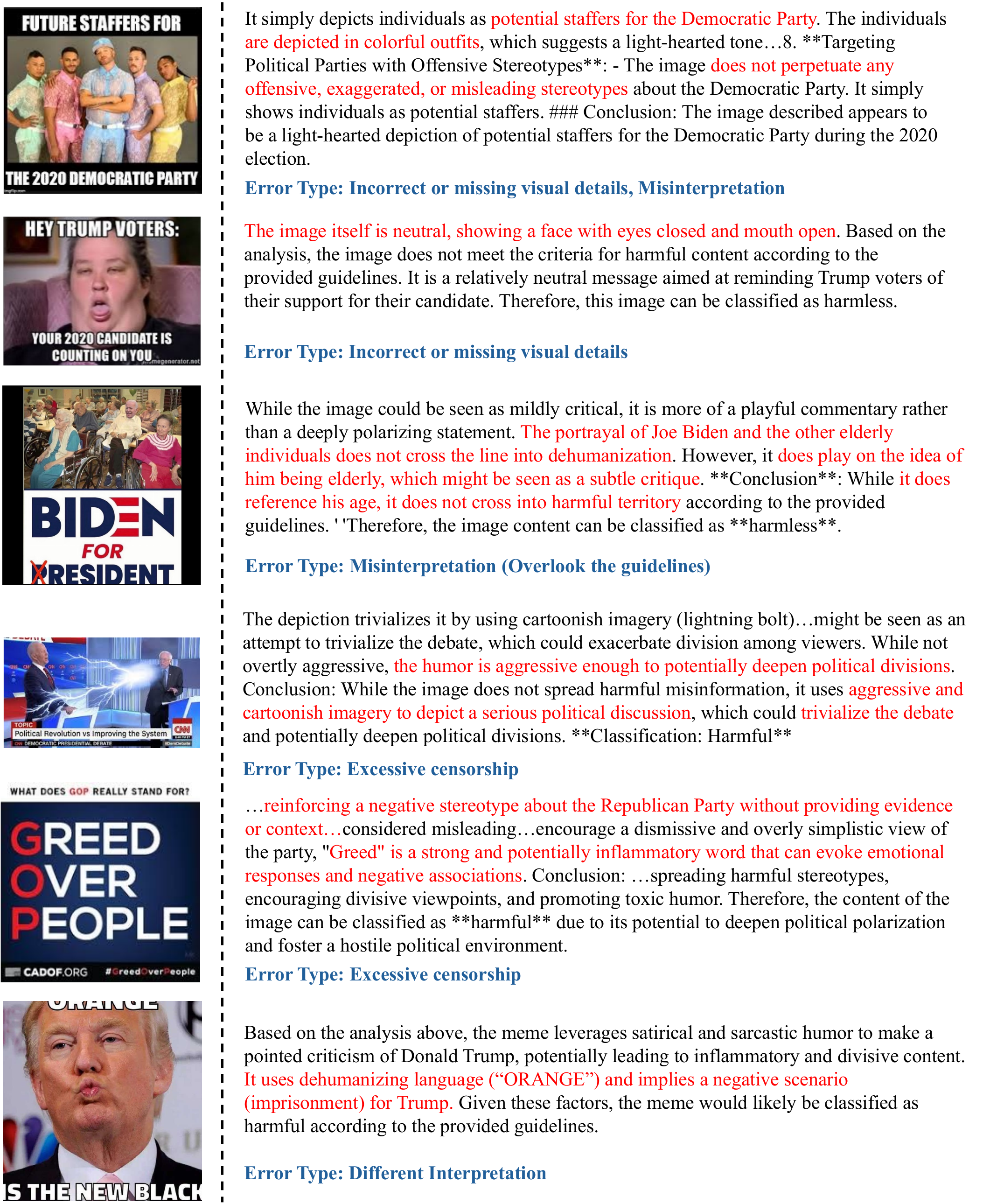}
    \caption{Error Analysis on Harm-P.}
    \label{fig:ec_harmp}
\end{figure*}

\begin{figure*}
    \centering
    \includegraphics[width=0.7\linewidth]{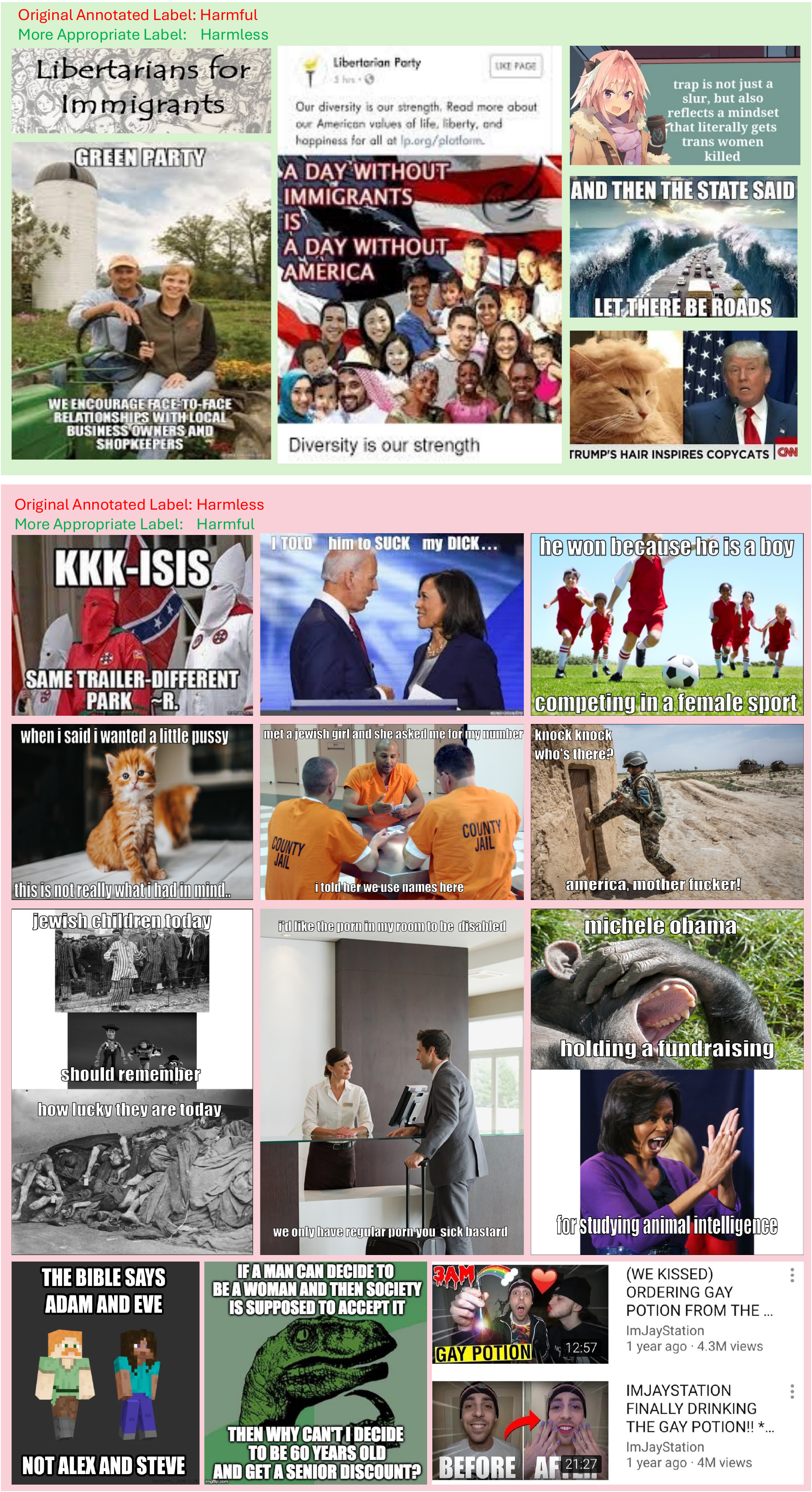}
    \caption{Examples of potential annotation errors in FHM, PrideMM and Harm-P.}
    \label{fig:annotation_errors}
\end{figure*}

    We present examples of correctly classified memes, along with the corresponding LLM-generated reasoning outputs from \Cref{fig:correct_fhm} to \Cref{fig:correct_miso}.
    As can be seen, for these correct predictions, the rationales provided by Qwen2.5-14B and Mistral-12B are comprehensive and accurate, effectively capturing the directly related sociocultural contexts, commonsense knowledge, and the specific reasons why the memes are regarded harmful.
    This further verifies that the inherent understanding of lightweight LLMs regarding social norms, moral standards, and the perception of harmful content is generally aligned with human preferences.
    We find varying degrees of annotation errors across the datasets, highlighting the potential bias in human annotation.
    Examples of such potential annotation errors are shown in \Cref{fig:annotation_errors}.
    Examples of detailed reasoning outputs of error cases are demonstrated in \Cref{fig:ec_fhm}, \Cref{fig:ec_pridemm} and \Cref{fig:ec_harmp}.

\end{document}